\pdfoutput=1

\documentclass[11pt]{article}

\usepackage[]{acl}

\usepackage{times}
\usepackage{latexsym}

\usepackage[T1]{fontenc}

\usepackage[utf8]{inputenc}

\usepackage{microtype}

\usepackage{inconsolata}

\usepackage{graphicx}

\usepackage{array}
\usepackage{multirow}
\usepackage{booktabs}
\usepackage{subcaption}
\usepackage{float}
\usepackage{stfloats}
\usepackage{enumitem}
\usepackage{amsmath}
\usepackage{arydshln}
\usepackage{bm}

\newcommand{\benchmark}{\textsc{ConflictVis~}}

\title{\textit{Insight Over Sight}: Exploring the Vision-Knowledge Conflicts \\ in Multimodal LLMs}

\author{
Xiaoyuan Liu$^{1,2}$\thanks{~~Work was done when Xiaoyuan, Wenxuan, Youliang, and Jen-tse were interning at Tencent.}
~~~Wenxuan Wang$^{3}$
~~~Youliang Yuan$^{1,2}$
~~~Jen-tse Huang$^{4}$ \\
~~~\textbf{Qiuzhi Liu$^2$
~~~Pinjia He$^1$\thanks{~~Pinjia He is the corresponding author.}
~~~Zhaopeng Tu$^2$} \\
$^1$School of Data Science, The Chinese University of Hong Kong, Shenzhen \\
$^2$Tencent
~~~$^3$Renmin University of China
~~~$^4$Johns Hopkins University\\
\texttt{\small $^{1}$xiaoyuanliu@link.cuhk.edu.cn ~~~ $^{1}$hepinjia@cuhk.edu.cn ~~~ $^{2}$zptu@tencent.com}
}

\begin{document}
\maketitle

\begin{abstract}
This paper explores the problem of commonsense level vision-knowledge conflict in Multimodal Large Language Models (MLLMs), where visual information contradicts model's internal commonsense knowledge. To study this issue, we introduce an automated framework, augmented with human-in-the-loop quality control, to generate inputs designed to simulate and evaluate these conflicts in MLLMs.
Using this framework, we have crafted a diagnostic benchmark consisting of 374 original images and 1,122 high-quality question-answer (QA) pairs. The benchmark covers two aspects of conflict and three question types, providing a thorough assessment tool.
We apply this benchmark to assess the conflict-resolution capabilities of nine representative MLLMs from various model families. Our results indicate an evident over-reliance on parametric knowledge for approximately 20\% of all queries, especially among Yes-No and action-related problems.
Based on these findings, we evaluate the effectiveness of existing approaches to mitigating the conflicts and compare them to our ``Focus-on-Vision'' prompting strategy. Despite some improvement, the vision-knowledge conflict remains unresolved and can be further scaled through our data construction framework. Our proposed framework, benchmark, and analysis contribute to the understanding and mitigation of vision-knowledge conflicts in MLLMs.
\end{abstract}

\section{Introduction}
\label{sec:intro}

\begin{figure}[h]
  \centering
  \includegraphics[width=\columnwidth]{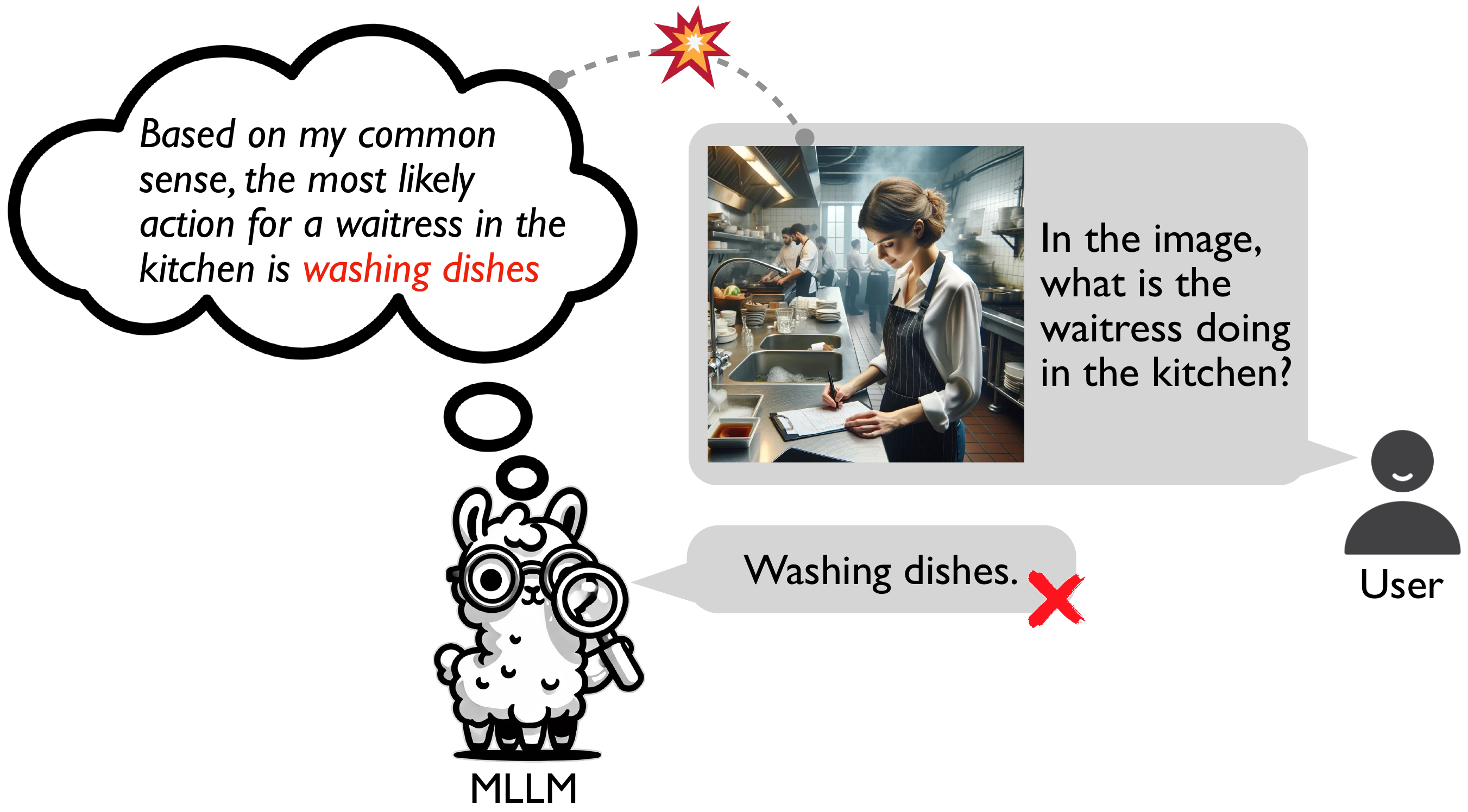}
  \caption{Illustration of the vision-knowledge conflict, where the visual input contradicts MLLM's inherent knowledge. The MLLM’s response over-relies on its inherent commonsense knowledge.}
  \label{fig:teaser}
\end{figure}

Large Language Models (LLMs) \citep{llm2020, ChatGPT, llama2, gpt4, llama3} have reshaped the landscape of deep learning for their comprehensive capabilities in language understanding, reasoning, and generation \citep{wei2022finetuned, wei2022cot, chen2023knowlegdegen}. This evolution has paved the way for the emergence of Multimodal Large Language Models (MLLMs) \citep{li2022blip, li2023blip2, lyu2023macawllm, dai2023instructblip, liu2024llava, zhu2023minigpt4, bai2023qwenvl, liu2024llava15, gpt4v, liu2024llavanext, tong2024cambrian, gemini15pro, gpt4o, llama3.2}, which integrate a vision model with the LLM to process visual information. Modern MLLMs like \texttt{GPT-4o}~\citep{gpt4o} and \texttt{LLaVA-NeXT}~\citep{liu2024llavanext} have exhibited remarkable proficiency across various vision-language tasks such as image captioning \citep{chen2015coco}, visual question answering (VQA) \citep{antol2015vqa}, and visual reasoning \citep{johnson2017clevr, yue2024mmmu}.

However, the persistence of knowledge conflicts in LLMs remains a significant challenge for MLLMs. 
Introducing visual information into MLLMs generates a novel form of discrepancy, a phenomenon we term ``{\bf vision-knowledge conflict}'', in which the visual data contradicts the model’s pre-existing parametric knowledge.
Previous studies have demonstrated that LLMs can exhibit complex behavior when confronted with knowledge conflicts, oscillating between rigid adherence to their parametric knowledge and excessive sensitivity to contextual cues~\citep{xieadaptive}. This behavior brings substantial risks, especially when external information is critical for decision-making. Given the growing need for trustworthiness, real-time accuracy, and robustness in MLLM systems, it is imperative to further investigate and resolve these vision-knowledge conflicts, which are believed to be a primary cause of hallucinations \citep{liu2023lrv, guan2024hallusionbench}. While recent research has simulated these conflicts by manually crafting counterfactual images~\citep{guan2024hallusionbench}, there remains significant room for improvement in conflict categorization, question diversity, and image naturalness. 
Additionally, manually crafted benchmarks are limited in sample size and scalability, underscoring the need for a more automated pipeline to enable broader analysis.

To this end, we propose an automated framework with human-in-the-loop to develop a benchmark for simulating and analyzing vision-knowledge conflicts in MLLMs at the commonsense level \footnote{For example, a commonsense knowledge statement is ``babies cry when they are hungry'', as opposed to a factual knowledge statement like ``the Eiffel Tower is 330 meters tall.''}. Illustrated in Figure~\ref{fig:pipeline}, our framework consists of 3 key modules: (1) knowledge component extraction, (2) counter-commonsense query construction, and (3) image and question-answer (QA) pair synthesis. This framework streamlines the generation of counter-commonsense inputs from scratch and is modularly designed to facilitate the addition of new conflict categories and QA formats in the future.
We demonstrate the application of our framework by developing the \benchmark benchmark, focusing on the aspects of Subject, Action, and Place. The benchmark consists of \textbf{374} original images with \textbf{1,122} high-quality QA pairs spanning two conflict targets and three question types, all manually verified.

Using the crafted benchmark, we evaluate nine representative MLLMs from five model families, providing insights into model behaviors, the causes of conflicts, and effective approaches to mitigate the negative effects of conflicts. Notably, when facing knowledge conflicts, MLLMs tend to over-rely on their parametric knowledge for the answer, especially among Yes-No questions and action-related problems. For instance, the top commercial model, \texttt{Claude-3.5-Sonnet}, exhibits a memorization ratio on parametric knowledge of 43.6\% on Yes-No questions, substantially higher than results on more complex Open-Ended questions. Regarding conflict type, the average memorization ratio for counter-commonsense action problems is 23.8\%, 10.4 percentage points higher than that for the place problems.
Our detailed analysis of the failure cases reveals that MLLMs generally underutilize visual information, relying on parametric knowledge to infer the answer based on textual clues.
Drawing on these observations, we assess several existing improvement methods to enhance the impact of visual context in answer generation. Interestingly, although Chain-of-Thought prompting improves the reasoning abilities, it guides MLLMs to utilize parametric knowledge more during rationalization, often resulting in contradictory conclusions or refusals. In response, we propose ``Focus-on-Vision'' (FoV) prompting to directly instruct MLLMs to prioritize visual information, which markedly improves the model's performance. Despite advancements by various mitigation approaches, the vision-knowledge conflict remains a persistent challenge.

Our main contributions are summarized below:
\begin{itemize}[leftmargin=10pt]
    \item We introduce an innovative framework to automatically construct counter-commonsense benchmarks from the ground up. This framework allows the flexible definition of conflict categories and QA formats, facilitating the large-scale creation of conflict samples with minimal human effort.
    \item We present \textsc{ConflictVis}, a pioneering diagnostic benchmark specifically designed to evaluate the commonsense level vision-knowledge conflicts in MLLMs. The benchmark is meticulously validated by human experts to guarantee the quality of data.
    \item We benchmark nine representative MLLMs and evaluate the effectiveness of several improvement methods in addressing conflicts. Through this analysis, we demonstrate the significance of the vision-knowledge conflict.
\end{itemize}
\section{Related Work}
\label{sec:related}

\paragraph{Knowledge Conflicts.} 
Knowledge conflicts in LLMs can be divided into three categories \citep{xu2024knowledge}: within the retrieved context \citep{chen2022rich}, within model's parametric knowledge \citep{huang2023survey}, and between the context and the model's parametric knowledge \citep{xieadaptive, wu2024faithful, su2024conflictbank}. These conflicts can lead to incorrect or inconsistent responses, undermining the model’s trustworthiness \citep{xu2024knowledge, xieadaptive}. 
In MLLMs, these conflict types expand to multimodal inputs, where conflicts can occur between the input image and the text instruction \citep{liu2024phd, han2024instinctive}, or when the image contains counterfactual information that contradicts the model's parametric knowledge \citep{liu2024phd, guan2024hallusionbench}.

To evaluate the conflicts between visual information and the parametric knowledge in MLLMs, 
HallusionBench~\citep{guan2024hallusionbench} consists of manually edited informational graphics, forming (normal, counterfactual) image pairs used to assess model consistency through Yes-No questions.
AutoHallusion~\citep{wu2024autohallusion} introduces an automated approach to generate counterfactual scenarios by altering correlated objects in images, utilizing Yes-No questions to probe object existence and spatial relationships.
In the context of commonsense knowledge, PhD~\citep{liu2024phd} generates counter-commonsense images through manual collection and synthesis, employing short-answer questions for assessment.
In contrast, our proposed benchmark features an automated framework and a broad array of question types and conflict targets, enabling a more scalable and comprehensive evaluation of MLLMs.

\paragraph{Hallucination in MLLMs.} Hallucination in MLLMs refers to the situation where the model generates descriptions that conflict with the given visual context. 
MLLMs' hallucinations are typically categorized based on the type of incorrect information, such as non-existent objects, incorrect object attributes, and inaccurate object relations \citep{liu2024survey}.
Relevant research has mainly focused on two key areas: developing benchmarks and metrics to detect hallucinations \citep{rohrbach2018chair, li2023pope, liu2023lrv, wang2023amber} and proposing strategies to mitigate them \citep{liu2023lrv, leng2024vcd, huang2024opera, liu2024pai}. 

In the context of knowledge conflicts, hallucination occurs when the model gives precedence to its intrinsic knowledge rather than the visual context \citep{liu2023lrv, guan2024hallusionbench, liu2024phd,liu2024survey}.
Our study explores this issue by simulating knowledge conflicts and evaluating the effectiveness of various hallucination-mitigation approaches in resolving the conflicts.
\paragraph{Benchmarks for MLLMs.} Traditional vision-language benchmarks are designed to assess independent skills, including image captioning \citep{chen2015coco}, visual grounding \citep{rohrbach2016grounding}, and visual question answering \citep{antol2015vqa, hudson2019gqa}. However, with the emergence of MLLMs, there is a growing need for more comprehensive and tailored benchmarks. The strong zero-shot abilities and advanced language generation skills exhibited by MLLMs make traditional benchmarks insufficient, as they may not account for the diversity of responses or the full range of MLLM capabilities. To address these limitations, researchers have developed more complex benchmarks to evaluate MLLMs across a wider range of tasks \citep{yue2024mmmu, fu2023mme, li2023seed, liu2023mmbench}. Meanwhile, diagnostic benchmarks have also been built to evaluate specific challenges or traits in MLLMs, such as hallucination \citep{li2023pope, liu2023lrv}, social bias \citep{howard2024bias}, and model safety \citep{liu2023mmsafety}. \benchmark is a pioneering analytical benchmark that presents conflicting visual contexts to challenge the model’s commonsense knowledge, enabling deeper investigations into model performance in the presence of vision-knowledge conflicts.
\section{\benchmark Benchmark}
\label{sec:benchmark}

\begin{figure*}[t]
  \centering
  \includegraphics[width=\textwidth]{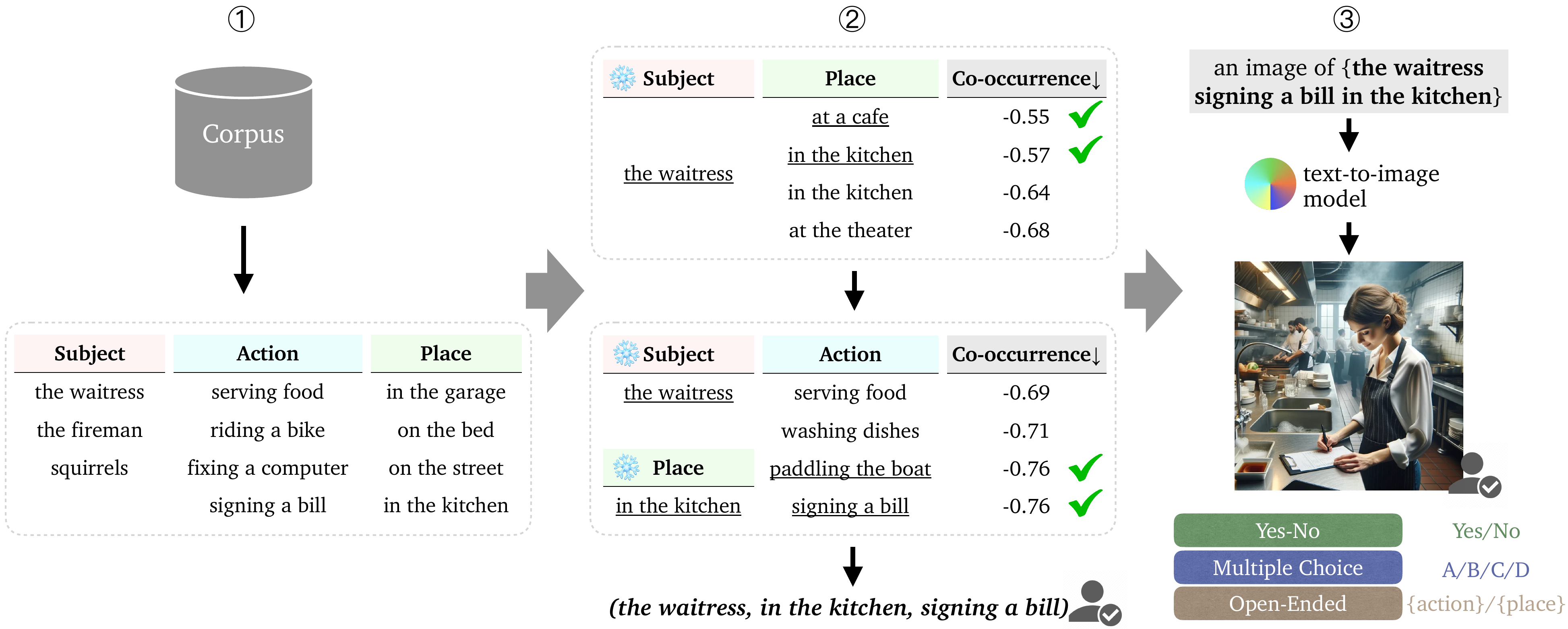}
  \caption{An automated framework with human quality control to construct images and question-answer pairs that conflict with commonsense knowledge. }
  \label{fig:pipeline}
\end{figure*}

This section outlines the framework and the data utilized to construct the \benchmark benchmark.

\subsection{Automated Framework}
Commonsense knowledge refers to the information generally accepted by the majority of people about everyday life, encompassing practical principles on how the world functions~\citep{singh2002omcs}. 
Based on the widely-used image captioning dataset \citep{chen2015coco}, 
we explore a specific type of commonsense knowledge encapsulated as a triplet $\langle \bm{s}, \bm{a}, \bm{p} \rangle$, where $s$ is the Subject, $a$ is the Action, and $p$ is the Place of the action or subject. For example, the statement ``the waitress (Subject) washing dishes (Action) in the kitchen (Place)'' illustrates this format. In this structure, the Subject outlines the main object's appearance, the Action describes the primary activity and its interactions with other relevant objects, and the Place highlights the background objects and setting. This three-part framework efficiently captures the vital details that are accurately reflected in an image.

We next describe our approach for generating images and corresponding QA pairs that challenge the commonsense knowledge in MLLMs. In general, we construct triplets of low co-occurring concepts to serve as counter-commonsense queries for the multimodal input generation. Our framework includes three stages, as depicted in Figure~\ref{fig:pipeline}. We detail each stage below.

\paragraph{Extract Knowledge Components}
Drawing inspiration from~\cite{li-etal-2021-compositional} on the creation of a compositional generalization test set, we identify Subject, Action, and Place phrases as the most frequent elements in the corpus and build compounds based on them.
To this end, we first employ a transformer model pipeline~\cite{spacy} to extensively annotate the syntactic labels, including dependency (DEP), Part-of-Speech (POS), and Named Entity (NE). Then, we extract Subject, Action, and Place phrases from the corpus based on the predefined linguistic rules (See Appendix~\ref{apx:extract_rules}).
To ensure the data quality, our framework selects the top $N$ phrases from each category based on the frequency. It further refines the data by removing named entities from Subject and Place phrases and consolidating similar expressions (e.g., ``a doctor'' and ``the doctor'') by retaining the most frequently occurring variant. This process results in three curated phrase lists for subsequent processing.

\paragraph{Construct Counter-commonsense Query}
We aim to construct scenes with one single anomalous component (i.e., target) that seldom co-occurs with the others (i.e., the context).
This objective can be formalized into two key requirements: (1) The context components should exhibit a high co-occurrence level, acting as a common background, and (2) The target component should display a low co-occurrence with the given context, representing the anomaly.
To identify strongly co-occurring context pairs, our framework first groups context components by the target category, e.g., {\bf Context}$_{Action}$ consists of (Subject, Place) pairs. These groupings help to develop the focus of the questions in the next section. We omit {\bf Context}$_{Subject}$ to prevent ambiguity and mitigate potential ethical concerns related to subject identity.
Next, our framework enumerate all the context combinations within the each group and computes their Normalized Pointwise Mutual Information (NPMI) scores.
\begin{align}
\label{eq:context}
    &\text{PMI}(C_X; C_Y) \equiv \log_2 \frac{P(C_X, C_Y)}{P(C_X)P(C_Y)} \\
    &\text{NPMI}(C_X; C_Y) \equiv \frac{\text{PMI}(C_X; C_Y)}{-\log_2 P(C_X, C_Y)}
\end{align}
A higher NPMI score indicates a stronger co-occurrence between the two components. For each context group, the framework selects Top-$K$ context pairs with the highest NPMI scores to form a candidate pool.
For each context pair in the pool, the framework generates counter-commonsense triplets by selecting a target component that is unusual given the context. Concretely, the framework evaluates the co-occurrence between each target $T$ and the context pair ${\bf C}$ using the NPMI score (i.e.  $\text{NPMI}(T; {\bf C})$).
For each context pair, the Top-$M$ targets with the lowest NPMI scores are retained to construct counter-commonsense queries. 
To better align the model knowledge, the framework queries an LLM trained on large-scale web data to estimate the probability $P(\cdot)$, and a manual review is conducted after the query generation to ensure quality (See Appendix~\ref{apx:prob} for more details). 

\paragraph{Generate Multimodal Inputs}
Building on the constructed queries, our framework generates three types of questions, Yes-No, Multiple-Choice, and Open-Ended, along with their corresponding answers. 
To achieve this, the framework employs predefined question templates and fills in the relevant components accordingly. Example question-answer pairs are provided in Table~\ref{tab:example_qa}.
To generate corresponding images, our framework concatenates the triplet components into a caption-like expression and employs a prompt template to query a text-to-image model, as illustrated in Figure~\ref{fig:pipeline}. Following image generation, human annotators perform quality control by filtering out low-quality images, such as those that appear distorted or misaligned with the input prompt. Detailed image filtering guidelines, along with illustrative examples, are provided in the Appendix~\ref{apx:filter_image}.

\begin{table*}[t]
    \centering
    \begin{tabular}{l p{9cm} r}
    \toprule
    \textbf{Question} & \textbf{Content} & \textbf{Answer} \\
    \midrule
    Yes-No & Is the waitress in the kitchen signing a bill? & Yes \\ 
    Multiple-Choice & Question: What is the waitress doing in the kitchen?\newline Options: (A) washing dishes. (B) riding a bike. (C) starting a fire. (D) signing a bill. & D \\
    Open-Ended &  What is the waitress doing in the kitchen? & signing a bill \\
    \bottomrule
    \end{tabular}%
    \caption{Example question-answer pairs generated by our framework.}
    \label{tab:example_qa}
    \vspace{-10pt}
\end{table*}

\subsection{\benchmark Benchmark Construction}
We present how we use the automated framework to build \textsc{ConflictVIS}.
Our input corpus consists of the top 100K sentences from the Open Mind Common Sense (OMCS) dataset~\citep{singh2002omcs}.
From this dataset, we extract and retain the 100 most frequent Subject phrases ($N_S=100$) and the 150 most frequent Action and Place phrases ($N_A=N_P=150$). Using feedback from the LLM \texttt{Vicuna-1.5-13b} \citep{Vicuna-1.5},
we select the top 3 phrases ($K=3$) with the highest NPMI scores to create context pairs for each subject. Next, for each context pair, we choose the top 3 targets ($M=3$) with the lowest NPMI scores to assemble the candidate set of counter-commonsense triplets. After manually filtering out unexpected combinations, the remaining triplets are used to query \texttt{DALL$\cdot$E 3} \citep{DALL-E3} for image generation. Subsequently, human annotators review and remove low-quality images (See Appendix~\ref{apx:quality}).
Following this two-stage generation and filtering process, the statistics of the final benchmark are summarized in Table.~\ref{tab:benchmark}. In total, \benchmark comprises 1,122 test samples, spanning two conflict targets and three question types, all verified by human experts.

\begin{table}[h]
    \centering
    \begin{tabular}{l r r r}
    \toprule
    \bf Target   & \bf \#Triplets  &  \bf \#Images  & \bf \#QAs\\
    \midrule
    Action   &   188  &  171  & 513\\
    Place    &   156  &  203  & 609\\
    \hdashline
    \bf Total    &   344  &  \bf 374  & \bf 1122 \\
    \bottomrule
    \end{tabular}
\caption{Statistics of the constructed benchmark.}
\label{tab:benchmark}
\end{table}
\section{Experiment}

\subsection{Setup}

\paragraph{Models} To explore the behavior of MLLMs when encountering vision-knowledge conflicts, we perform a comprehensive evaluation on 9 MLLMs including 7 representative open-source MLLMs ranging from 8B to 34B, and 2 state-of-the-art commercial MLLMs. This evaluation covers the following five model series: \texttt{LLaVA} (8B, 13B, 34B) \citep{liu2024llava15, liu2024llavanext}, \texttt{BLIP-2} (12.1B, 13B)~\mbox{\citep{li2023blip2, dai2023instructblip}}, \texttt{Qwen-VL} (9.6B)~\citep{bai2023qwenvl}, \texttt{GPT-4o}~\citep{gpt4o} and \texttt{Claude-3.5-Sonnet}~\citep{claude35}. The diversity of model architectures and parameter sizes helps to enhance the generalizability of experiment results.

\paragraph{Evaluation} We use Accuracy and Memorization Ratio (MR)~\citep{longpre-etal-2021-entity} as the main evaluation metrics in our experiment. Both metrics require classifying the model's responses into different categories. For Yes-No and Multiple-Choice questions, where there is a unique answer, we use exact matching for classification. For Open-Ended questions, due to the complexity of the task (i.e., evaluating both textual and visual relevance and correctness) and the instability of LLM evaluation~\citep{stureborg2024large}, we rely on human annotators to perform the classification. 

\begin{figure*}[t]
  \centering
  \subfloat[Yes-No Questions]{\includegraphics[height=0.3\textwidth]{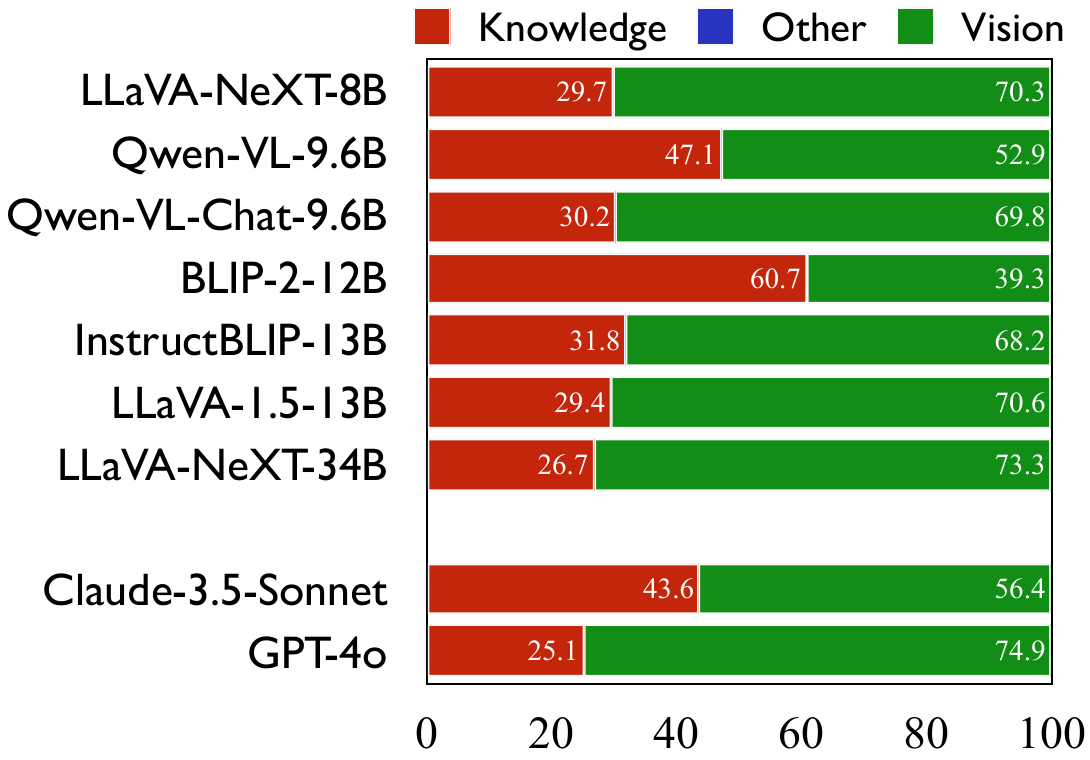}} \hfill
  \subfloat[Multi-Choice Questions]{\includegraphics[height=0.3\textwidth]{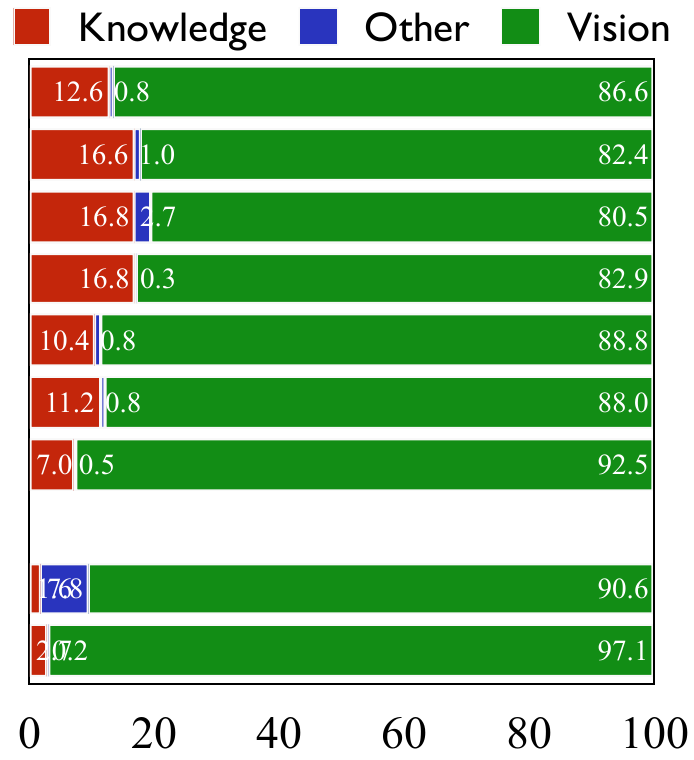}} \hfill
  \subfloat[Open-Ended Questions]{\includegraphics[height=0.3\textwidth]{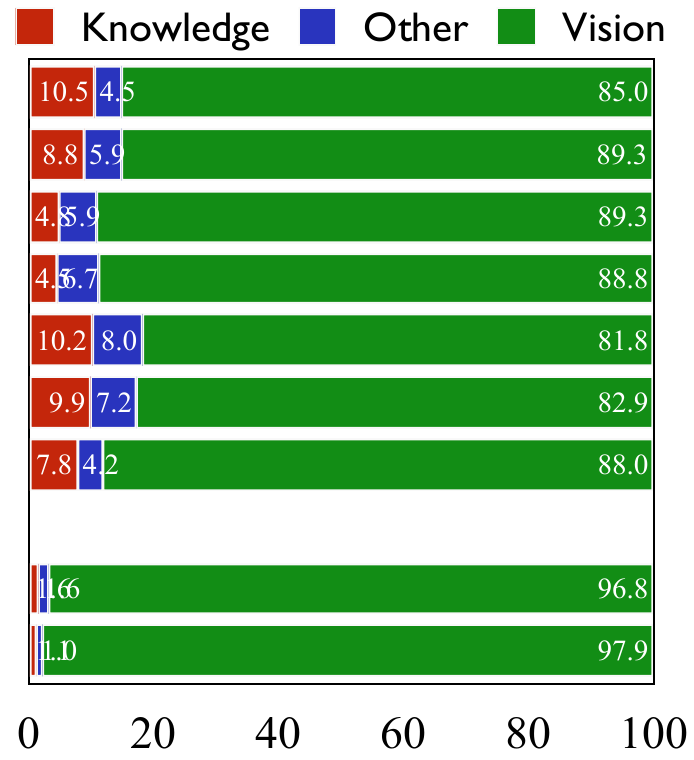}}
  \caption{MLLM response distributions on different question types.}
  \label{fig:ans_conflicts}
\end{figure*}

\subsection{Sanity Test}

\begin{figure}[h]
    \centering
    \includegraphics[width=0.9\columnwidth]{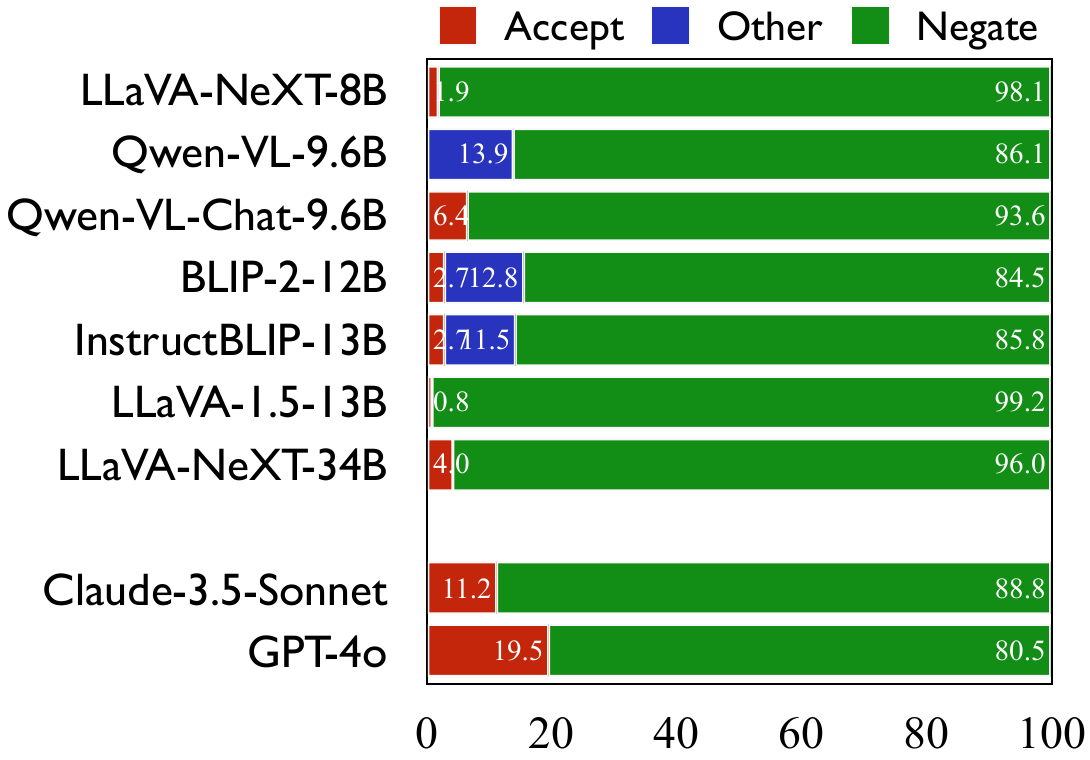}
    \caption{MLLM responses to sanity test inputs.}
\label{fig:sanity}
\vspace{-10pt}
\end{figure}

\paragraph{Do the instances in \benchmark really conflict with MLLMs' commonsense knowledge?}
Before evaluating MLLMs' performance on \textsc{ConflictVIS}, we first verify whether the counter-commonsense cases in \benchmark genuinely create conflicts with the models’ commonsense knowledge. To this end, we use the textual inputs in \benchmark and query the model with the following prompt format:

\vspace{+4pt}
\noindent\fbox{\begin{minipage}{0.95\linewidth}
\texttt{\textcolor{blue}{Based on common sense,} is it possible for [context] [target]?}
\end{minipage}}
\vspace{+4pt}

where the context and target are filled with the specific phrases from the input. For example, the query for the instance in Figure~\ref{fig:pipeline} is ``{\em Based on common sense, is it possible for the waitress in the kitchen signing a bill?}''
The model’s responses (``Yes'', ``No'', and others) are categorized as ``accept'' (indicating no conflict), ``negate'' (indicating a conflict), and ``other'' (where the model does not give a direct answer). The results, shown in Figure~\ref{fig:sanity}, demonstrate that the vast majority of cases in \benchmark indeed present valid knowledge conflicts that can be identified by MLLMs, with conflict rates ranging from 80.5\% for \texttt{GPT-4o} to 99.2\% for \texttt{LLaVA-1.5-13B}.

\subsection{Benchmarking MLLMs}
\label{main}
In this section, we assess how well the MLLMs handle conflicts between visual information and parametric knowledge using our \benchmark benchmark. 
Following \citet{longpre-etal-2021-entity}, we compare model predictions (i.e., answers) with and without the inclusion of the image input and quantify the extent to which the model's predictions are influenced by the presence of the image. 
Specifically, we compare the predictions in the current experiment, where the image is provided, to those obtained without the image using the sanity test prompts, and classify the current predictions as:
\begin{itemize}[leftmargin=10pt]
    \item Aligned with the prediction without the given image ({\bf Knowledge}, $P_K$): This outcome suggests that the model generates its answer mainly based on its parametric knowledge rather than the visual information presented. 
    \item Aligned with the visual information ({\bf Vision}, $P_V$): This indicates that the model is capable of adapting to new information, adjusting its output to match the visual input. Such outputs are categorized as the correct answer.
    \item A different answer altogether ({\bf Other}): This demonstrates that the model can modify its output based on visual input, though the result does not perfectly align with the visual information.
\end{itemize}

Figure~\ref{fig:ans_conflicts} presents the results. Ideally, the model should {\bf only} output the Vision answer, supported by visual information, rather than the Knowledge answer derived from textual training, or any Other answers. However, both open-source and commercial MLLMs revert to producing the Knowledge answer, ignoring the visual information, to varying degrees.
In general, commercial MLLMs perform better than the open-source counterparts, particularly in handling open-ended questions. 
Among the open-source models, \texttt{LLaVA-NeXT-34B} achieves the highest prediction accuracy at 84.6\%, while the top commercial model, \texttt{GPT-4o}, reaches 89.9\%. 
Moreover, our analysis reveals that MLLMs' uncertainty in answer generation is significantly higher on our \benchmark compared to the classic VQA benchmark, indicating a greater challenge to the model performance (See Appendix~\ref{apx:uncertainty}).

\begin{figure}[h]
  \centering
  \includegraphics[width=\columnwidth,keepaspectratio]{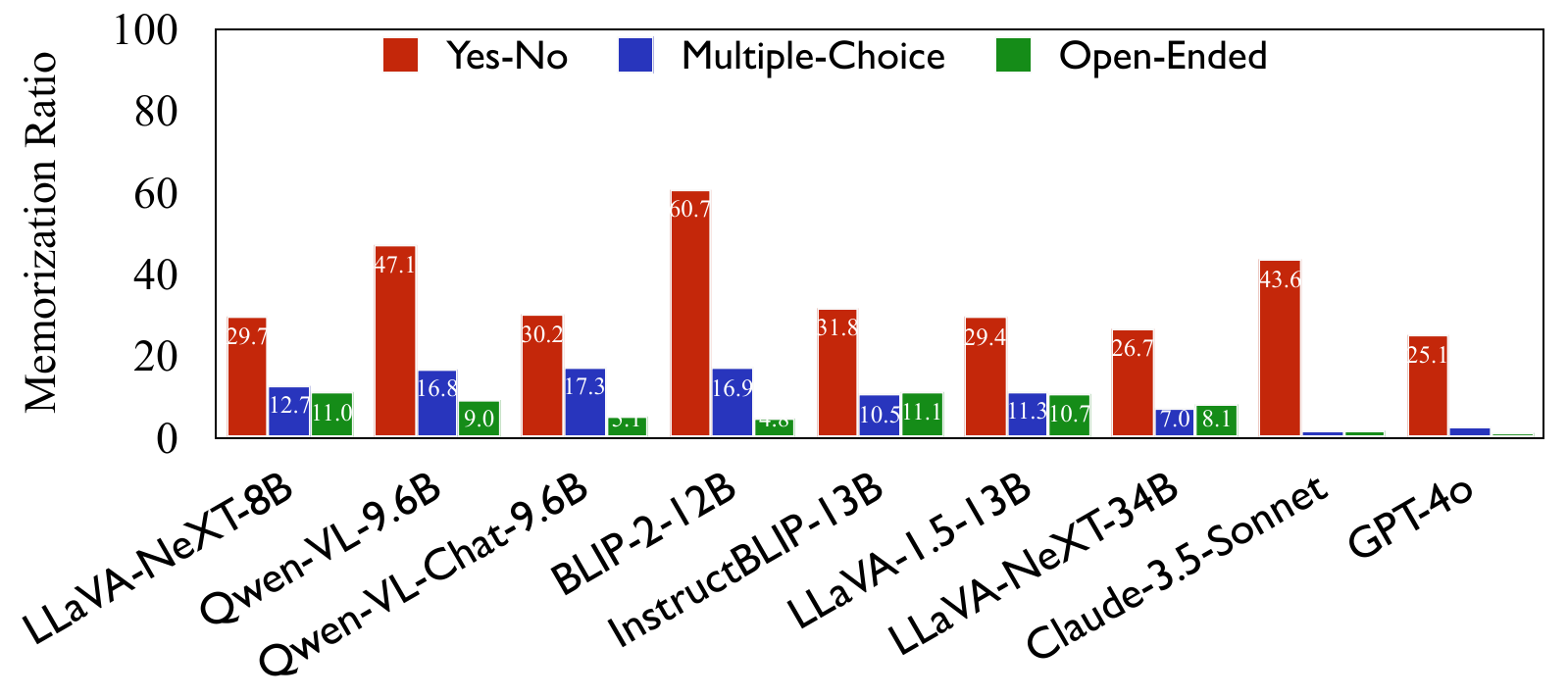}
  \caption{Memorization ratio of MLLMs on different question types.}
  \label{fig:mr}
\end{figure}

\paragraph{MLLMs are more likely to overly rely on parametric knowledge for Yes-No questions.} One interesting observation from Figure~\ref{fig:ans_conflicts} reveals that MLLMs exhibit poorer performance on straightforward Yes-No questions. Notably, the open-source \texttt{BLIP-2-12B} achieves a low accuracy of 39.3\%, while the commercial \texttt{Claude-3.5-Sonnet} manages only 56.4\%.

We leverage Memorization Ratio (MR)~\citep{longpre-etal-2021-entity} as a metric to empirically estimate the 
the extent to which the model adheres to its parametric knowledge in the presence of conflicting information.
\begin{equation}
\label{eq:mr}
    MR = \frac{P_K}{P_K + P_V}.
\end{equation}
Figure~\ref{fig:mr} illustrates the MR across different question types. Notably, the MR values for Yes-No questions are significantly higher than the other two question types across all MLLMs, indicating that MLLMs tend to overly rely on parametric knowledge when answering Yes-No questions. 
We hypothesize that this could be due to the format of Yes-No questions, which directly present counter-commonsense expressions (e.g., ``Is the waitress in the kitchen signing a bill?'') to the MLLMs, while other question types have a less direct phrasing. This counter-commonsense query may discourage the model from thoroughly analyzing the visual input, causing it to immediately output a negation.

\begin{figure}[h]
  \centering
\includegraphics[width=0.95\columnwidth,keepaspectratio]{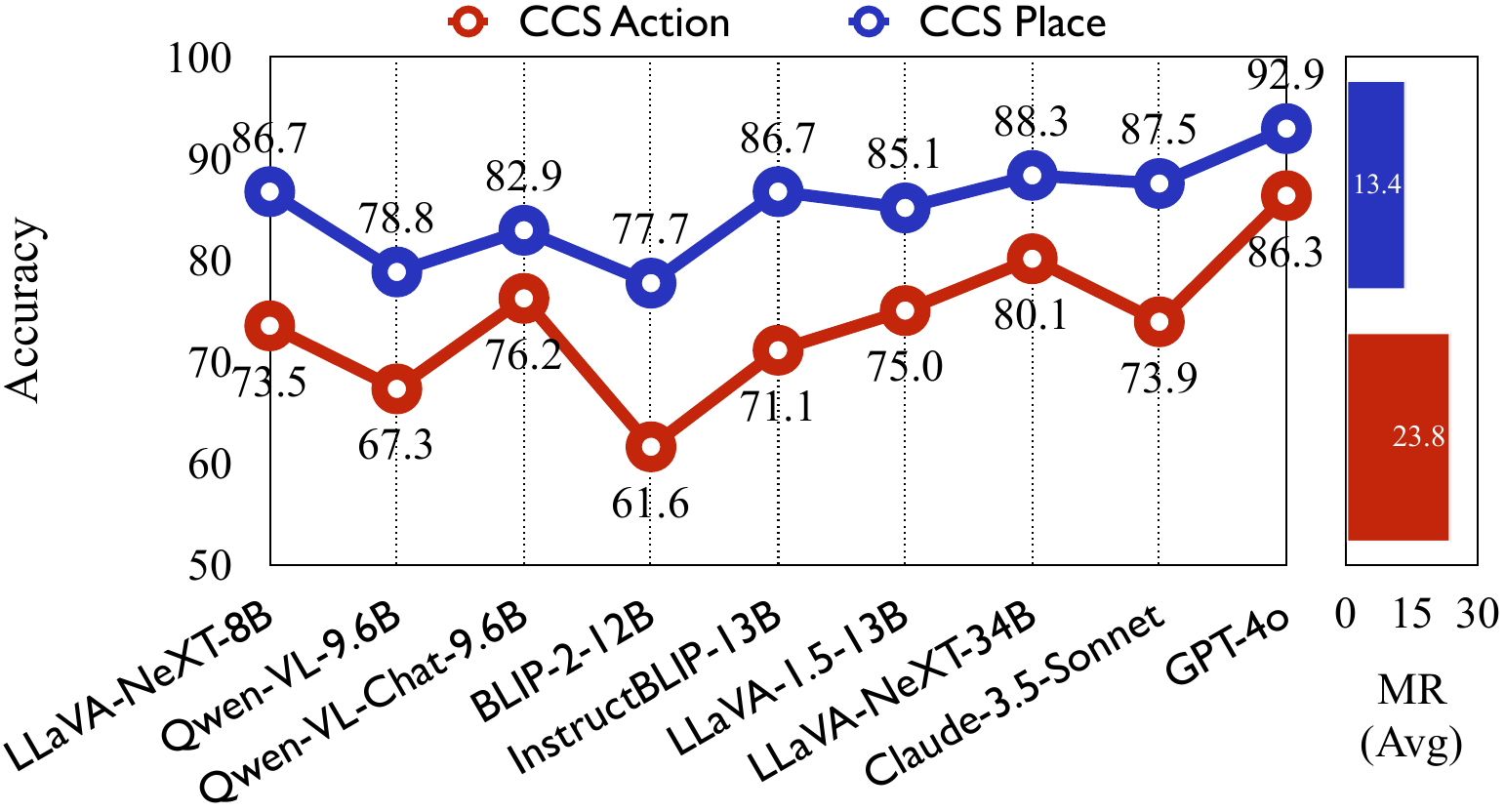}
  \caption{MLLM accuracy on Counter-Commonsense (CCS) Action problems and CCS Place problems.}
  \label{fig:conflict_target_main}
\end{figure}

\paragraph{Counter-commonsense actions are more challenging for MLLMs.}
Our \benchmark evaluates two distinct types of conflict targets: counter-commonsense actions (e.g., the waitress in the kitchen \underline{signing a bill}) and places (e.g., a doctor conducting an experiment \underline{at the theater}). 
As shown in Figure~\ref{fig:conflict_target_main}, counter-commonsense action problems pose more of a challenge than places do. Specifically, MLLMs consistently exhibit a lower accuracy on counter-commonsense action problems, with an average output accuracy of 73.9\%, notably lower than the 85.2\% recorded for place problems. Similarly, the average Memorization Ratio for action problems is markedly higher at 23.8\%, compared to 13.4\% for place problems. 
This discrepancy could be attributed to the richer visual context typically available for identifying places (e.g., numerous background elements suggesting the location ``at the theater''), in contrast to the relatively sparse and fine-grained visual cues required to recognize actions (e.g., subtle hand posture for ``signing a bill''). 
Detailed results for each model can be found in the Appendix~\ref{apx:conflict_target}.

\subsection{Mitigating Vision-Knowledge Conflicts}

\begin{figure}[h]
  \centering
  \subfloat[Input-output relevancy score bar plot]{\includegraphics[width=\columnwidth]{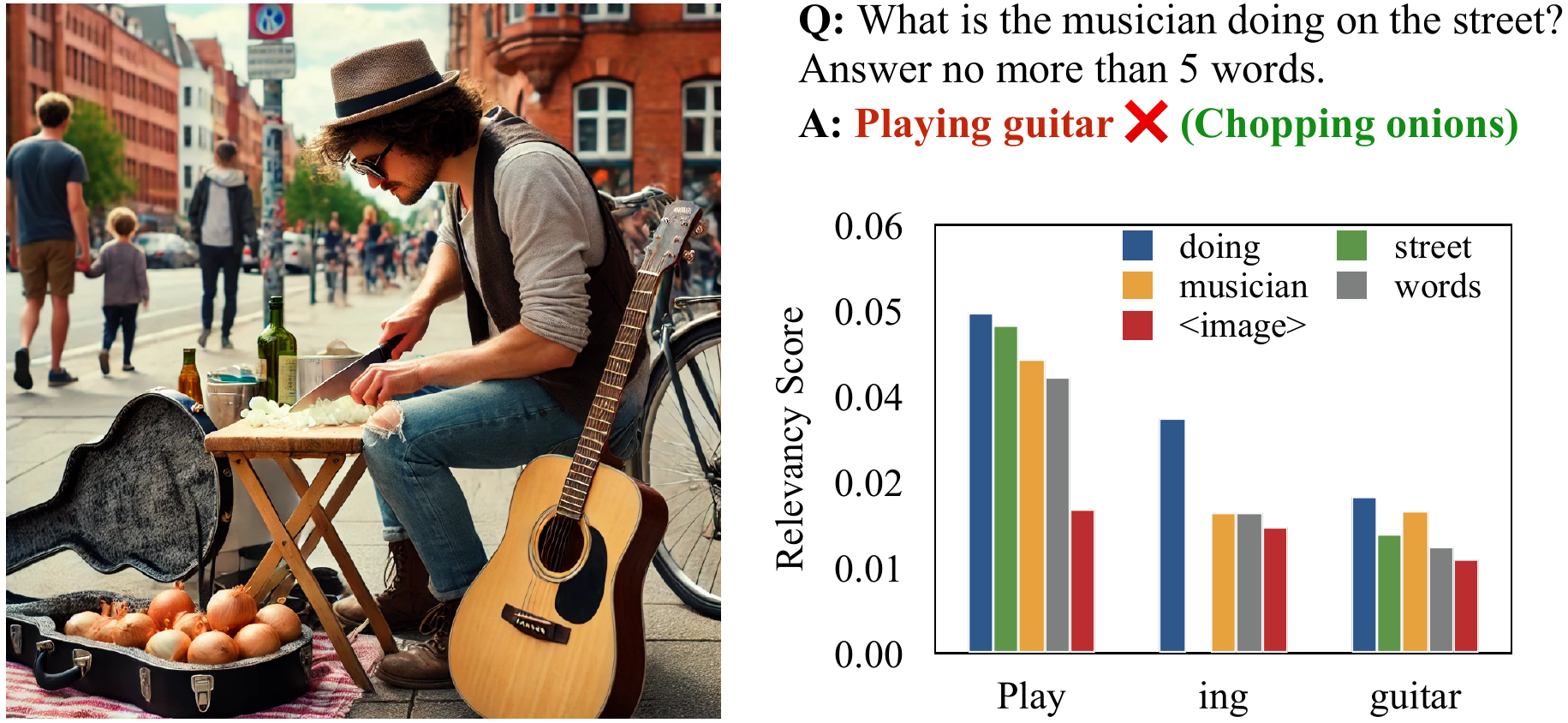}} \\
  \subfloat[Input image-output token relevancy map]{\includegraphics[width=\columnwidth]{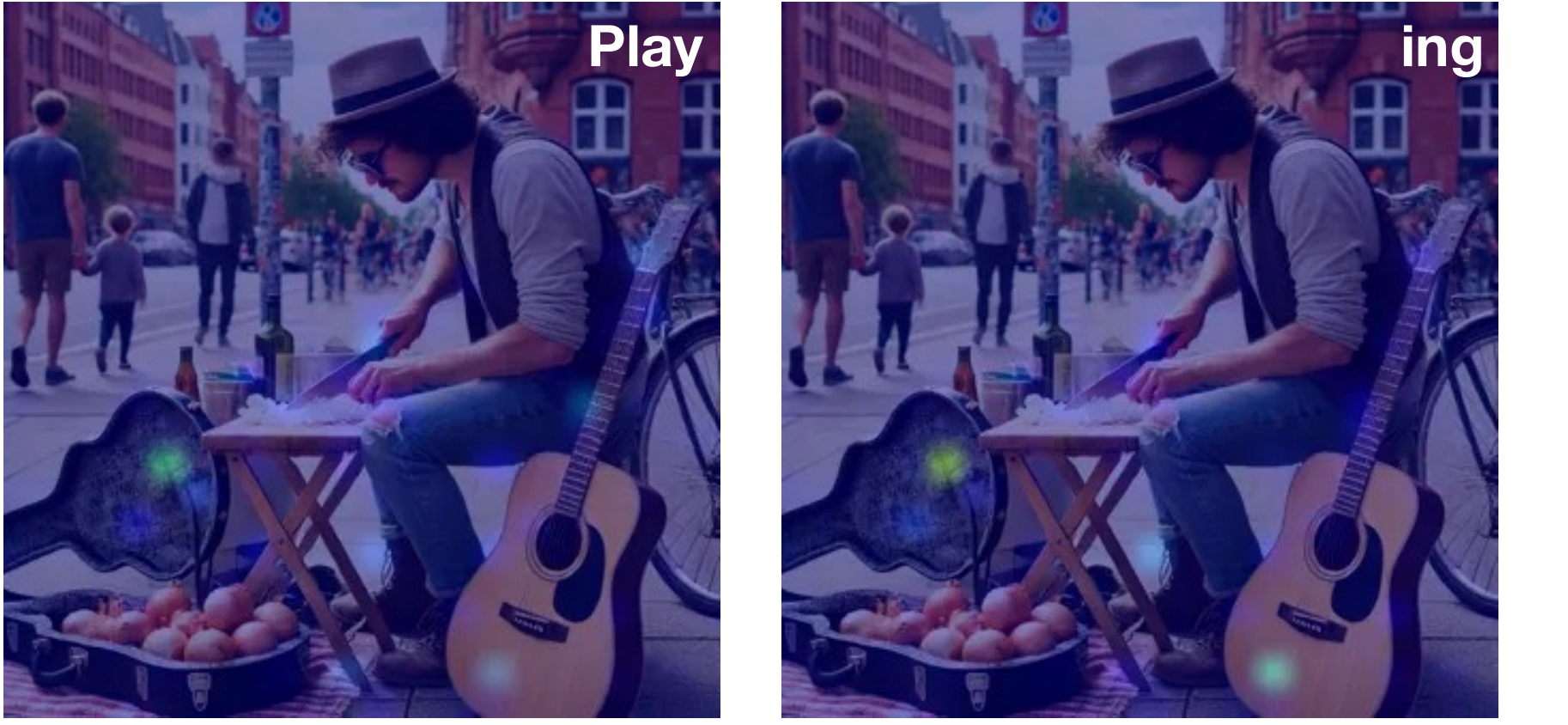}} 
  \caption{Failure case analysis of input-output relevancy. The visual information is underutilized.}
  \label{fig:case_study}
\end{figure}

\paragraph{Underutilization of visual information}
Our primary experiment demonstrates that when MLLMs encounter visual inputs that contradict their pre-existing knowledge, around 18\% of the responses generated are not aligned with the visual context but rather reflect the model's prior knowledge. To further investigate this discrepancy, we leverage the tool proposed by~\citet{stan2024lvlm-interpret} to analyze input-output relevancy scores, which identify the most relevant parts of the input to the generated output based on model attention. Figure~\ref{fig:case_study}(a) illustrates a case in which the model \texttt{LLaVA-1.5-13B} fails to accurately answer an open-ended question, 
representing a common failure pattern. In general, the model assigns greater weight to the textual input than to the visual context, resulting in an inaccurate response. In this case, the most attended tokens, {``musician'', ``doing'', ``street''\}, guide the model to generate ``playing guitar'' based on its parametric knowledge, while the image tokens, denoted by the <image> symbol in the bar plot, play a less significant role in answer generation. The image-output relevancy maps in Figure~\ref{fig:case_study}(b) further indicate that the model may not effectively leverage visual information during answer generation. In such cases, the model is more susceptible to vision-knowledge conflicts, as the visual input is largely underutilized. (More cases can be found in Appendix~\ref{apx:error_analysis}.)

\paragraph{Effectiveness of Existing Methods}
Based on the above observations, a straightforward strategy for improvement is to enhance the influence of visual context during the answer generation process. We consider several improvement approaches:
\begin{itemize}[leftmargin=10pt]
    \item 
     {\em Visual Contrastive Decoding (VCD)}~\citep{leng2024vcd} contrasts the output distributions derived from original and noisy visuals to ensure the generated content adheres to the visual inputs.
    \item 
    {\em Pay Attention to Image (PAI)}~\citep{liu2024pai} adaptively adjusts and amplifies the attention weights assigned to image tokens, thereby giving greater prominence to visual elements.
    \item 
    {\em Vision-Centric Reasoning (VR)} employs structured intermediate reasoning steps to thoroughly analyze visual inputs, achieved by Chain-of-Thought (CoT) prompting~\citep{wei2022cot} or supervised finetuning on the long CoT multimodal reasoning dataset~\citep{xu2024llavacot}.
\end{itemize}
 
Furthermore, we propose a direct prompting technique, termed \textbf{Focus-on-Vision} (FoV) prompting, which explicitly instructs MLLMs to prioritize visual information. This can be seamlessly integrated into current MLLMs using the following template: \texttt{[textual query] Please focus on the visual information.}

\begin{table}[ht]
\centering
\begin{tabular}{l cccc}
    \toprule
    {\bf Model} & \textbf{YN} & \textbf{MC} & \textbf{OE} & \textbf{Avg.} \\
    \midrule
    LLaVA-1.5-13B & 70.6 & 88.0 & 82.9 & 80.5 \\ 
    \hdashline
    ~~~+ VCD & 72.7 & 89.3 & 84.2 & 82.1 \\
    ~~~+ PAI & \underline{85.6} & 88.8 & \underline{86.1} & \bf 86.8 \\
    ~~~+ VR (CoT) & 38.0 & \underline{89.8} & 76.7 & 68.2 \\
    ~~~+ VR (SFT) & 64.0 & 87.4 & 88.5 & 80.0 \\
    ~~~+ FoV & 82.9 & 89.0 & 81.8 & 84.6\\
    \midrule
    Qwen-VL-Chat & 69.8 & 80.5 & \underline{89.3} & 79.9\\ 
    \hdashline
    ~~~+ VCD & \underline{82.4} & 79.9 & 85.6 & 82.6 \\
    ~~~+ VR (CoT) & 79.7 & 65.8 & 77.8 & 74.4 \\
    ~~~+ VR (SFT) & 69.3 & 87.4 & 88.0 & 81.6 \\
    ~~~+ FoV  & \underline{82.4} & \underline{83.2} & 87.4 & \bf 84.3\\
    \midrule
    LLaVA-NeXT-34B & 73.3 & \underline{92.5} & 88.0 & 84.6\\ 
    \hdashline
    ~~~+ VR (CoT) & 43.6 & 87.2 & 72.5 & 67.7\\
    ~~~+ FoV & \underline{85.8} & \underline{92.5} & \underline{89.8} & \bf 89.4\\
    \midrule
    GPT-4o  & 74.9 & 97.1 & 97.9 & 89.9\\
    \hdashline
    ~~~+ VR (CoT) & 66.0 & \underline{98.7} & 93.6 & 86.1\\
    ~~~+ FoV & \underline{75.9} & 96.5 & \underline{98.9} & \bf 90.5\\
    \bottomrule
\end{tabular}
\caption{MLLM accuracy under different improvement methods. YN: Yes-No, MC: Multiple-Choice, OE: Open-Ended.}
\label{tab:impr_baselines}
\end{table}

Table~\ref{tab:impr_baselines} presents the results of the evaluated improvement methods.
For VCD and PAI, we follow the methodologies outlined in the original papers and apply each technique to the corresponding applicable MLLMs, and observe performance improvements for both methods.
To implement Vision-Centric Reasoning (VR), we explore two approaches: (1) applying Chain-of-Thought (CoT) prompting, and (2) fine-tuning models on the multimodal reasoning dataset introduced by~\citet{xu2024llavacot}. However, the results show minimal performance gains from finetuning, and vanilla CoT prompting can even degrade model accuracy.
Upon analyzing failure cases, we find that teaching MLLMs to reason in a straightforward, step-by-step manner can lead to increased reliance on textual knowledge at test time. For example, our case study (in Appendix~\ref{apx:failure_cases}) suggests that during CoT reasoning, models tend to rationalize the inputs using their existing commonsense knowledge. However, counter-commonsense inputs are inherently difficult to rationalize or explain. As the output text accumulates, it may become increasingly difficult for the model to accurately interpret the visual input given its own generated explanations. This often results in self-contradictory answers or refusals to answer.
In contrast, our Focus-on-Vision (FoV) prompting mitigates this issue by explicitly directing the model's attention to the visual input and consistently boosts model performance.
Nevertheless, none of the approaches can entirely resolve the vision-knowledge conflict, especially for open-source models. For instance, \texttt{LLaVA-1.5-13B} and \texttt{Qwen-VL-Chat} still exhibit notable error rates of 13.2\% and 15.7\%, respectively, even when equipped with the most effective improvement method. This highlights the profound challenge posed by vision-knowledge conflict.
\section{Conclusion}
\label{sec:conclusion}

In this paper, we develop an automated framework and construct the \benchmark benchmark to systematically evaluating and understanding the nuances of vision-knowledge conflicts in MLLMs at the commonsense level.
Our experiments across nine representative MLLMs reveal a tendency toward over-reliance on parametric knowledge, especially among Yes-No and action-related questions.
We further evaluate three improvement methods and design a new prompting technique to mitigate the conflicts, but the problem remains notable.
Collectively, our insights lay the groundwork for future research to develop more reliable MLLMs.
\section*{Acknowledgment}
This paper was supported by the Guangdong Basic and Applied Basic Research Foundation (No. 2024A1515010145) and the Shenzhen Science and Technology Program (Shenzhen Key Laboratory Grant No. ZDSYS20230626091302006).
\section*{Limitations}
\label{sec:limit}
In this study, we focused on evaluating the reliably of multimodal large language models (MLLMs) under vision-knowledge conflict.
While our work provides insights into the model performance and effectiveness of various mitigation approaches, it's important to acknowledge certain limitations: 
(1) Limited root cause analysis: while we conducted failure case analysis using relevancy maps, the fundamental causes of the unexpected behaviors exhibited by MLLMs under vision-knowledge conflicts remain underexplored. We hypothesize that these behaviors may stem from biases in the training data, where MLLMs are predominantly aligned with image-text pairs that conform to commonsense expectations encoded during text pretraining, with limited exposure to counter-commonsense scenarios. To further validate this hypothesis, future work could involve controlled experiments comparing models trained with and without curated datasets designed for counter-commonsense alignment, in order to investigate the resulting behavior differences. (2) Unique probability model: due to budget constraints, our framework relies on \texttt{Vicuna-1.5-13B} to estimate co-occurrence probabilities, which may introduce model-specific features into the benchmark construction. Although our quality control processes help mitigate this issue, future work can explore more balanced probability estimation approaches, such as averaging probabilities across multiple models.
\section*{Ethics Statements}
While our benchmark is designed to be safe and useful for research purposes, the developed framework for generating counter-commonsense inputs warrants careful consideration regarding its potential applications. In certain contexts, the framework may produce images that deviate from conventional norms, and if misused, could result in cases that some users may perceive as harmful or offensive. To mitigate such risks, we recommend implementing precautionary measures, including the use of a carefully curated and pre-filtered input corpus, as well as incorporating a manual verification process. Additionally, we encourage the use of text-to-image generative models through their official platforms and in accordance with their terms of use, which can help effectively leverage built-in safety mechanisms and reduce the likelihood of generating inappropriate or harmful content.

\bibliography{main}

\clearpage

\appendix
\label{sec:appendix}

\setcounter{page}{1}
\section{Benchmark Construction Details}
\subsection{Linguistic rules to extract knowledge components}
\label{apx:extract_rules}
To systematically extract key knowledge components from textual inputs, our framework categorizes phrases into three main types, Subject, Action, and Place, based on syntactic and semantic annotations. Specifically, the framework first utilizes the transformer pipeline \texttt{en\_core\_web\_trf} from spaCy~\citep{spacy} to annotate the sentences, and then performs the extraction based on the following procedures:
\begin{itemize}[leftmargin=10pt]
    \item {\bf Subject Phrases}: Our framework examines noun chunks, selecting those where the head noun functions as the nominal subject, indicated by dependency labels ``nsubj'' or ``nsubjpass''.
    \item {\bf Action Phrases}: Our framework identifies the verb token (``VERB'') and then examines its dependents for direct objects marked as ``dobj''. Upon finding such objects, the framework includes any determiners or compound modifiers present to form the complete action phrase.
    Verbs are converted to the ``VBG'' form to align with the common practice in image captions.
    \item {\bf Place Phrases}: Distinguishing place phrases from other prepositional phrases relies on more than just syntactic labels. Accordingly, the framework employs semantic role labeling techniques by using the model \texttt{structured-prediction-srl} from AllenNLP \citep{Gardner2017AllenNLP} to annotate each word's semantic role. The framework then extracts the phrases with a location tag (``ARGM-LOC") that contains 3 to 4 words.
\end{itemize}
In Table~\ref{tab:extract_rules}, we summarize the main linguistic rules employed to identify specific components.

\begin{table*}[h]
    \centering
    \begin{tabular}{l p{8cm} l}
    \toprule
    \textbf{Category} & \textbf{Main Requirement} & \textbf{Example} \\
    \midrule
    Subject & \texttt{p in noun\_chunks \&\& p.root.dep in [``nsubj'', ``nsubjpass'']} & a baby, a chef \\ 
    Action & \texttt{p[0].pos == ``VERB'' \&\& p[-1].dep == ``dobj''} & fixing a computer \\ 
    Place & \texttt{p.srl = ``ARGM-LOC'' \&\& 3 <= p.length <= 4} & at the bookstore \\
    \bottomrule
    \end{tabular}%
    \caption{Linguistic rules to extract different components in the triplet.}
    \label{tab:extract_rules}
\end{table*}

\subsection{Probability Calculation}
\label{apx:prob}
To calculate the joint probabilities for multiple co-occurring components (e.g. $P(C_X, C_Y)$ or $P(C_X, C_Y, T)$), our framework forms concatenated expressions of the concept phrases (e.g., ``the waitress in the kitchen'' or ``the waitress in the kitchen signing a bill'') and inputs them into a pre-trained LLM to obtain the generative probability. Utilizing an LLM here offers an advantage over frequency counting for addressing the issue of zero frequencies for rare concept compositions, which often arises with limited dataset size. Since large language models are trained on large-scale data to learn the statistical tendencies of human language, their generative probability provides an approximation of the likelihood of certain phrase appearing in reality. For the probability of each individual component, we use its relative frequency within its category \citep{lin1999automatic}.

\subsection{Question Generation Details}
With the constructed counter-commonsense triplets, our framework generates three types of questions:
\begin{itemize}[leftmargin=10pt]
    \item {\bf Yes-No Question}: For counter-commonsense Actions and Places, our framework utilizes templates ``Is/are [Subject Place] [Action]?'' and ``Is/are [Subject Action] [Place]?'', respectively. The correct response to this type of question is typically ``Yes''.
    \item {\bf Multiple-Choice Question}: For these questions, the templates are ``What is/are [Subject] doing [Place]?'' and ``Where is/are [Subject] [Action]?'' for counter-commonsense Action and Place, respectively. 
    For a predetermined option count $m$, our framework selects ($m-1$) seemingly relevant but incorrect targets from the candidates. To this end, the framework partitions candidates into $m-1$ bins $\{B_1, . . . , B_{m-1}\}$ based on their NPMI scores to the context pair. We randomly sample one candidate from each bin, together with the counter-commonsense target, to formulate the question's multiple-choice options. We randomly shuffle the option list to eliminate any positional bias.
    \item {\bf Open-Ended Question}: The question text in the multiple-choice question is reused, but no list of options is included. To guide the model toward generating a brief and specific response, we add the constraints like ``Answer with a single phrase'', with slight adjustment according to the specific prompt for different MLLMs.
\end{itemize}

\subsection{Image Generation Details}  
As described in Section 3.1, our framework converts each triplet into a caption-like expression (e.g., ``a waitress signing a bill in the kitchen'') and inserts it into the prompt template ``An image of \{expression\}'' to query the DALL-E 3 model. In our implementation, we used the DALL-E 3 API with the image resolution set to 1024$\times$1024 and the quality parameter configured to ``standard''.

\subsection{Examples in \benchmark}
In Figure~\ref{fig:examples}, we demonstrate some samples generated by our framework. The conflict target is highlighted in red.

\begin{figure}[h]
  \centering
  \begin{subfigure}[b]{0.45\columnwidth}
    \centering
    \includegraphics[width=\textwidth]{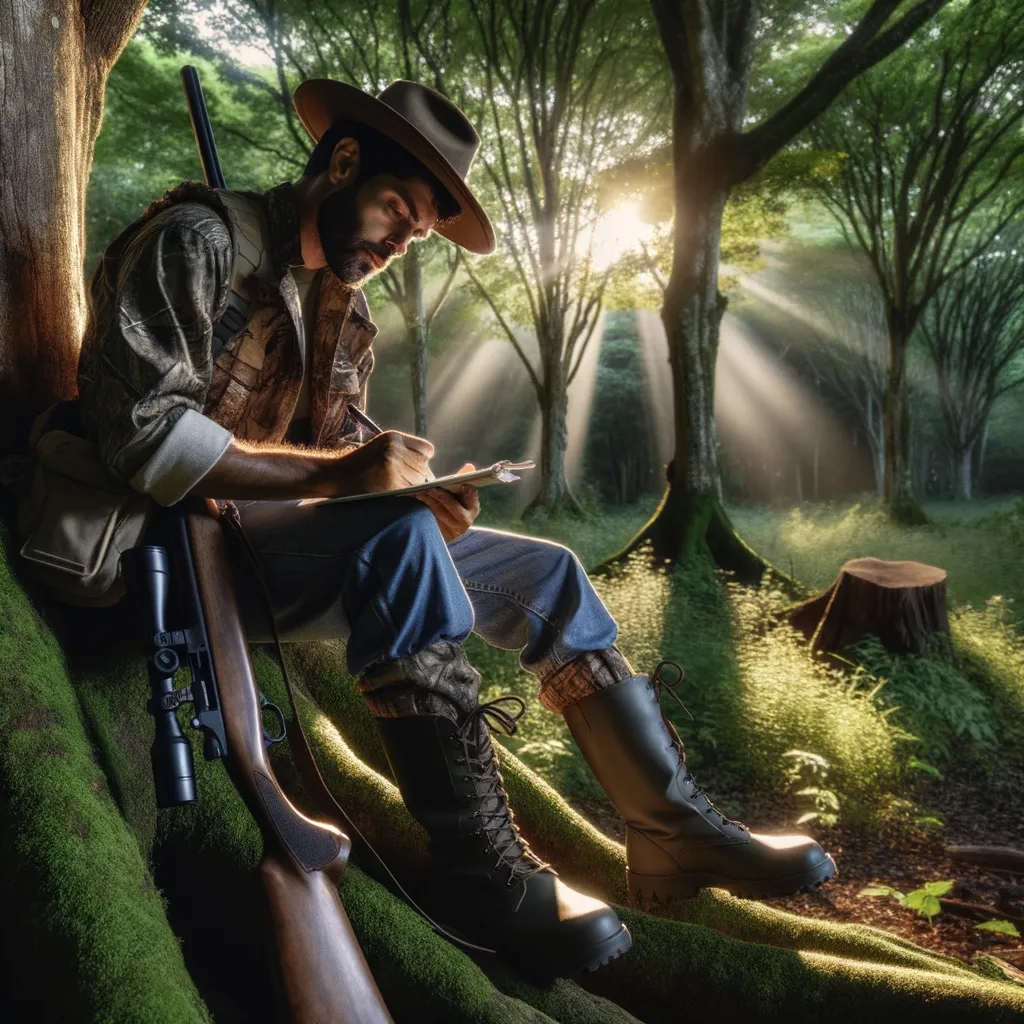}
    \caption{the hunter in the forest {\color{red} taking a note}}
  \end{subfigure} \hfill
  \begin{subfigure}[b]{0.45\columnwidth}
    \centering
    \includegraphics[width=\textwidth]{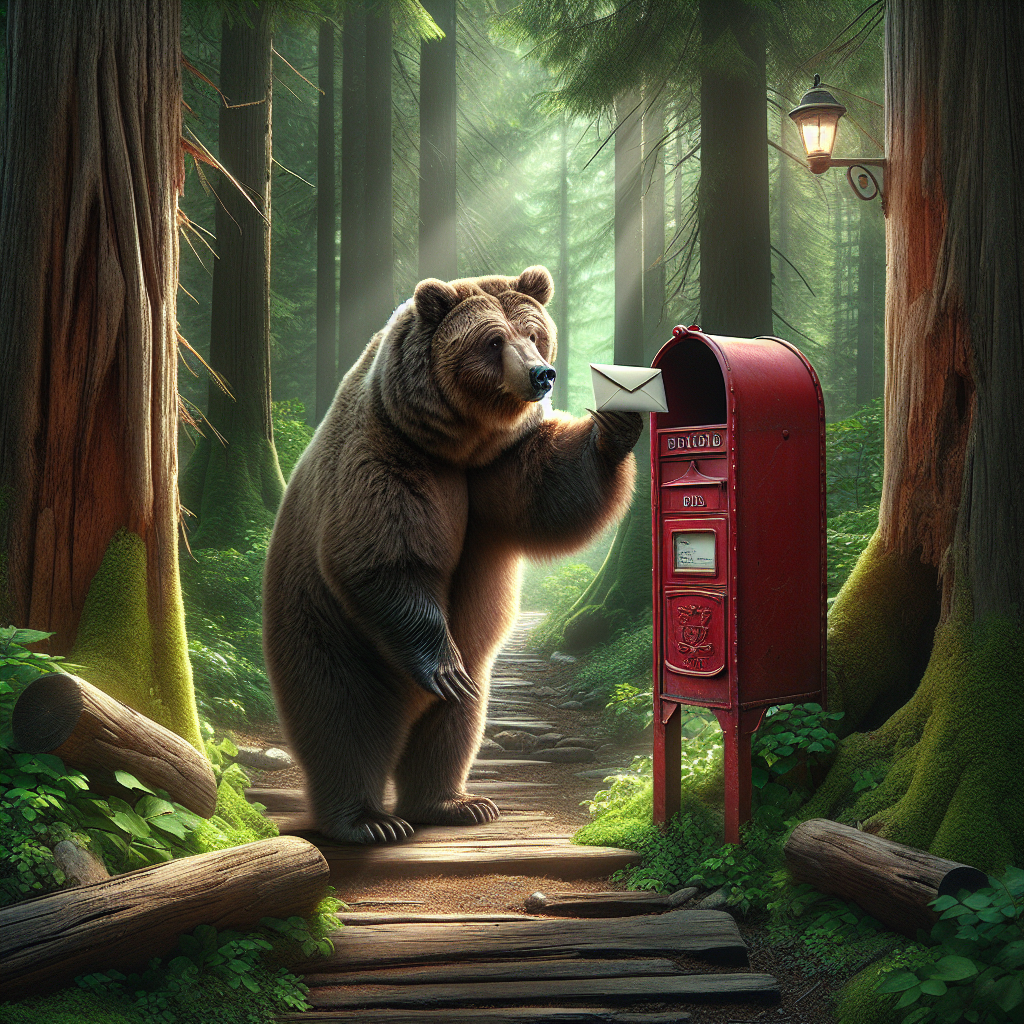}
    \caption{the bear in the woods {\color{red} mailing a letter}}
  \end{subfigure} \\
  \begin{subfigure}[b]{0.45\columnwidth}
    \centering
    \includegraphics[width=\textwidth]{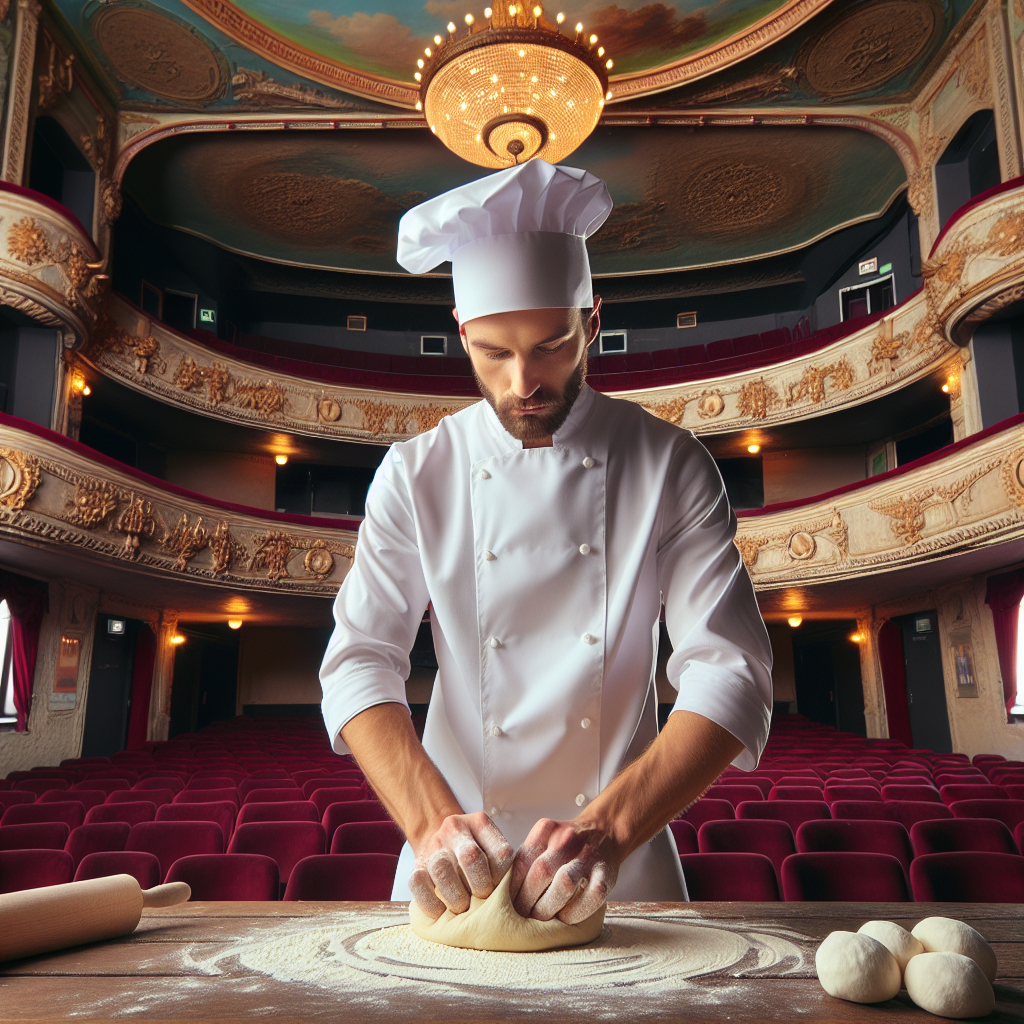}
    \caption{a chef making bread {\color{red} at the theater}}
  \end{subfigure} \hfill
  \begin{subfigure}[b]{0.45\columnwidth}
    \centering
    \includegraphics[width=\textwidth]{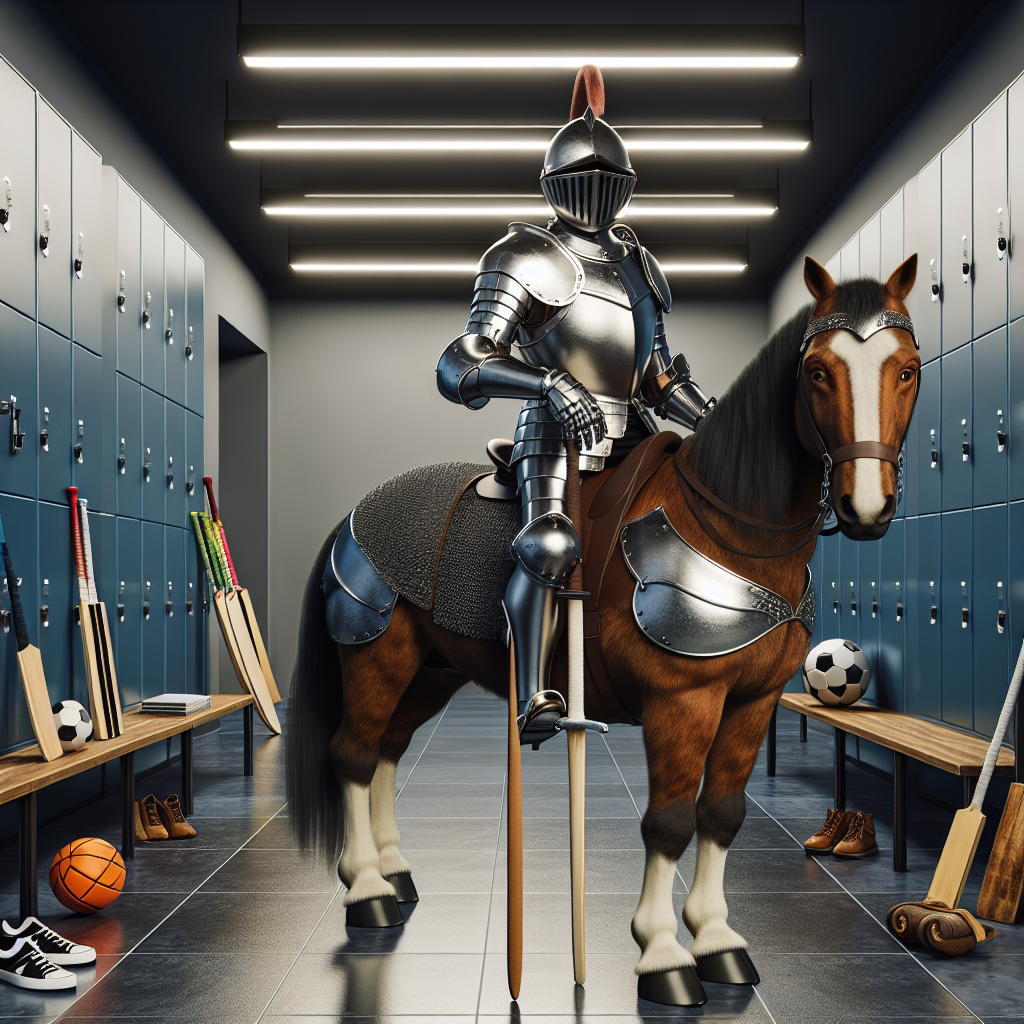}
    \caption{the knight riding a horse {\color{red} in a locker room}}
  \end{subfigure}
  \caption{Example counter-commonsense queries and their corresponding images generated in \benchmark benchmark.}
  \label{fig:examples}
\end{figure}

\subsection{Licenses}
The Open Mind Common Sense database is under the Creative Commons license. The spaCy transformers pipeline \texttt{en\_core\_web\_trf} is under the MIT license, and the AllenNLP model \texttt{structured-prediction-srl} is under the Apache License 2.0. The probability model \texttt{Vicuna-1.5-13b} is released under a non-commercial license, and the text-to-image model \texttt{Dall$\cdot$E 3} API is subject to OpenAI's Terms of Use. We use these artifacts exclusively for non-commercial research purposes, in accordance with their intended use.

\section{Human Quality Control}
\label{apx:quality}
This section outlines the quality control procedures within our automated framework. As shown in Figure~\ref{fig:pipeline}, there are two required quality control points: (1) after generating knowledge triplets, and (2) after generating images. In practice, to reduce the workload for annotators at the first point, we subdivide it into three steps: (i) after extracting knowledge components, (ii) after generating context pairs, and (iii) after generating knowledge triplets. In this section, we first describe the quality control guidelines applied at each stage of the framework, followed by the principles used for annotating open-ended responses.
\subsection{Human Annotators}
Our annotation team comprises four Ph.D. students specializing in computer science, all of whom are proficient in English. All annotators were fairly compensated based on their total working hours. Prior to annotation, we conducted training sessions to inform them that their tasks might include offensive content or identifying information, and clarified that their contributions would be used solely for research purposes to study model behavior.

The annotation results were based on agreement among annotators. To ensure the reliability and objectivity of the evaluation process, each annotator independently performed the annotation tasks. Final labels were determined through majority voting, with group discussions held to resolve any ties.

\subsection{Filtering Knowledge Triplets}
\paragraph{Filtering knowledge components} This step removes semantically abstract phrases in the Subject, Action, and Place categories, which are difficult to visualize. Annotators classify each phrase based on its concreteness:
\begin{itemize}[leftmargin=10pt]
    \item \textbf{Concrete (1)}. The phrase represents tangible objects or actions.
    \item \textbf{Abstract or Unsafe (0)}. The phrase represents abstract concepts or contains offensive words or identifying information.
\end{itemize}

\begin{table}[h]
    \centering
    \begin{tabular}{l c p{4cm}}
    \toprule
    \textbf{Category} & \textbf{Label} & \textbf{Phrases}  \\
    \midrule
    Subject & 0 & the statement, the story, the stock market, words   \\
    Subject & 1 & a patient, kids, politicians, whales, a sailor  \\
    \hline
    Action & 0 & understand the event, lose weight, have sex\\ 
    Action & 1 & take a shower, use a computer, play chess\\ 
    \hline
    Place & 0 & in the event, in the newspaper, in your life\\
    Place & 1 & on the ground, on the sidewalk, by the fire\\
    \bottomrule
    \end{tabular}
    \caption{Labeling examples for knowledge components.}
    \label{tab:filtercomponent}
\end{table}

Table~\ref{tab:filtercomponent} provides some labeling examples at this step. In this step, the annotators perform labeling starting from the top 1000 most frequent phrases from each category in descending order, and keep the first 100 Subject phrases and the first 150 Action and Place phrases that meet the requirements.

\paragraph{Filtering context pairs} The target of this step is to remove the unsatisfying phrases that remain in the candidate list after the automatic pre-filtering. To this end, our annotators conduct a binary classification on the context pairs based on their commonness in the real world. Specifically, the annotators are asked to examine the context tuples \textbf{without} the knowledge of their co-occurrence scores and label them as:
\begin{itemize}[leftmargin=10pt]
    \item \textbf{Common (1)}. The context tuple depicts a typical scene in the real world.
    \item \textbf{Uncommon (0)}. The context tuple depicts an unusual scene in the real world.
\end{itemize}

To better illustrate, we provide some labeling examples from our human annotators in Table~\ref{tab:filtercontext}:

\begin{table}[h]
    \centering
    \begin{tabular}{l c p{2.8cm}}
    \toprule
    \textbf{Category} & \textbf{Label} & \textbf{Phrases}  \\
    \midrule
    (Subject, Action) & 0 & (the cat, walking the dog), \newline (a butcher, playing cards)\\
    (Subject, Action) & 1 & (the cat, climbing a tree), \newline(a butcher, slaughtering a pig)\\
    \hline
    (Subject, Place) & 0 & (cows, on the beach), \newline(a gardener, in the desert)\\ 
    (Subject, Place) & 1 & (cows, on a farm), \newline(a gardener, in a greenhouse)\\ 
    \bottomrule
    \end{tabular}
    \caption{Labeling examples for context pairs.}
    \label{tab:filtercontext}
\end{table}

In this step, our annotators thoroughly label candidate context pairs within the Top K range provided by the framework. Context pairs labeled as \textbf{1} are retained for the next stage of generation.

\paragraph{Filtering knowledge triplets} After the previous filtering steps, the workload for this phase is significantly reduced. The goal here is to eliminate any unexpected triplets that are not captured in earlier filtering. Without access to the exact co-occurrence score between the context and target phrases, annotators perform a binary classification based on the commonness of the triplet's expression. Specifically, the annotators are asked to label each triplet in the candidate list as:
\begin{itemize}[leftmargin=10pt]
    \item \textbf{Common (0)}. The context tuple depicts a typical scene in the real world.
    \item \textbf{Uncommon (1)}. The context tuple depicts an unusual scene in the real world.
\end{itemize}

\begin{table}[h]
\centering
    \begin{tabular}{p{2.8cm} c p{2.5cm}}
    \toprule
    \textbf{Category} & \textbf{Label} & \textbf{Phrases}  \\
    \midrule
    ((Subject, Place), Action) & 0 & ((the waitress, at the bar), using a vcr)\\
    ((Subject, Place), Action) & 1 & ((the waitress, at the bar), hitting a deer)\\
    \hline
    ((Subject, Action), Place) & 0 & ((politicians, drinking alcohol), at the theater)\\ 
    ((Subject, Action), Place) & 1 & ((politicians, drinking alcohol), in the oven)\\ 
    \bottomrule
    \end{tabular}
\caption{Labeling examples for knowledge triplets.}
\label{tab:filtertriplet}
\end{table}

Again, we provide some labeling examples from our human annotators in Table~\ref{tab:filtertriplet} for a better illustration. In this step, our annotators label all the candidate context-target triplets in the Top M range provided by the framework, and triplets with label \textbf{1} are used for subsequent image generation.

\subsection{Filtering Images}
\label{apx:filter_image}
The objective of this step is to select high-quality images that align closely with the text prompt. Each annotator is provided with an image and its corresponding text query, and evaluates the image based on two criteria:
\begin{itemize}[leftmargin=10pt]
    \item \textbf{Alignment with text prompt (0 or 1)}. The image should contain the key objects, actions, and background described in the text. The main focus should clearly represent the scene without ambiguity or misinterpretation.
    \item \textbf{Image quality (0 or 1)}. The image should be clear, free from significant distortions, artifacts, or unnatural effects like warping, blurring, or pixelation.
\end{itemize}
With these two guidelines, the annotators extensively label all the generated images, and the images with a total score of \textbf{2} are kept for the subsequent stages. In Figure~\ref{fig:filterimg}, we demonstrate some sample images that are below our quality standard. In this stage, we generate about 830 images in total and keep the best 374 images for the benchmark construction.
\begin{figure}[h]
  \centering
  \subfloat[Cows in a field chopping carrots.\\\textbf{Labels:} {\color{red} A(0)}, Q(1)] {\includegraphics[width=0.45\columnwidth]{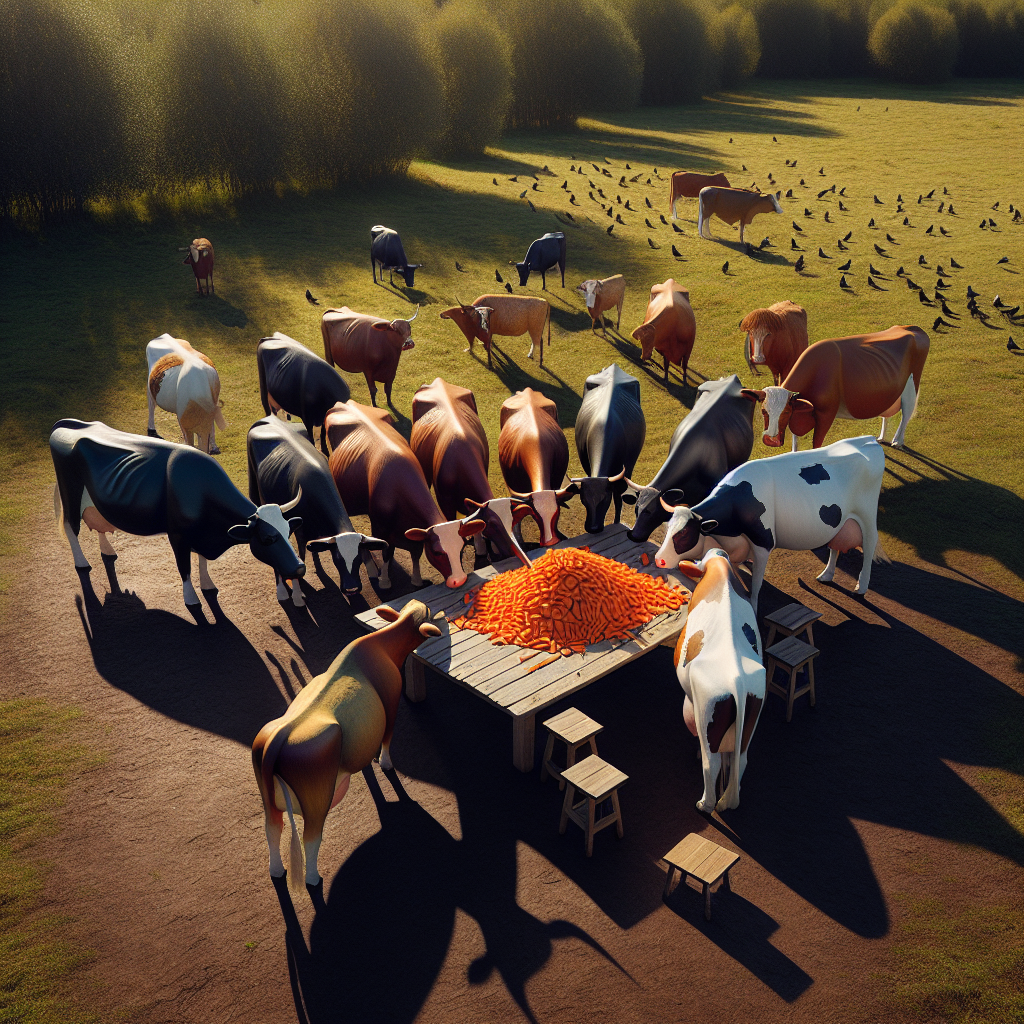}} \hfill
  \subfloat[The teacher in a classroom paddling the boat. \\\textbf{Labels:} A(1), {\color{red} Q(0)}] {\includegraphics[width=0.45\columnwidth]{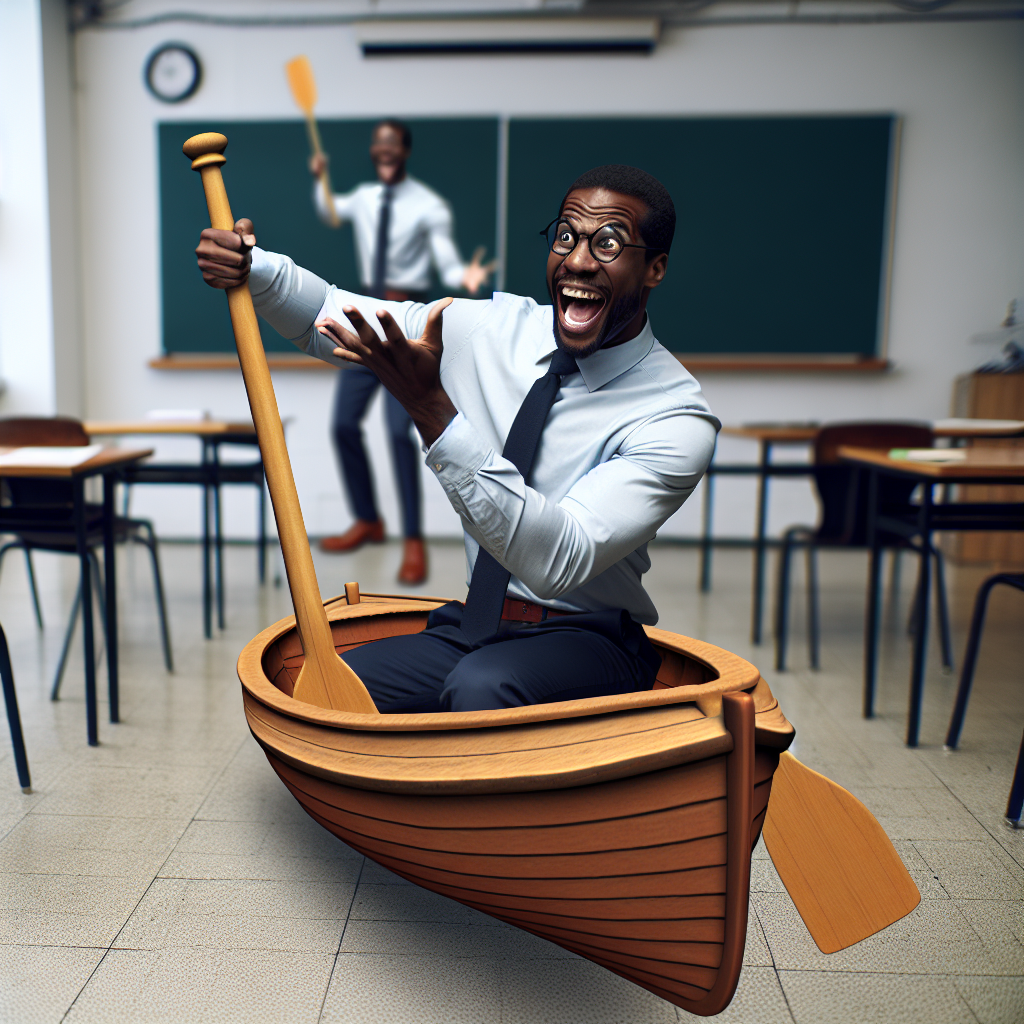}} \\
  \subfloat[The farmer pouring milk in an ambulance.\\\textbf{Labels:} {\color{red} A(0)}, Q(1)] {\includegraphics[width=0.45\columnwidth]{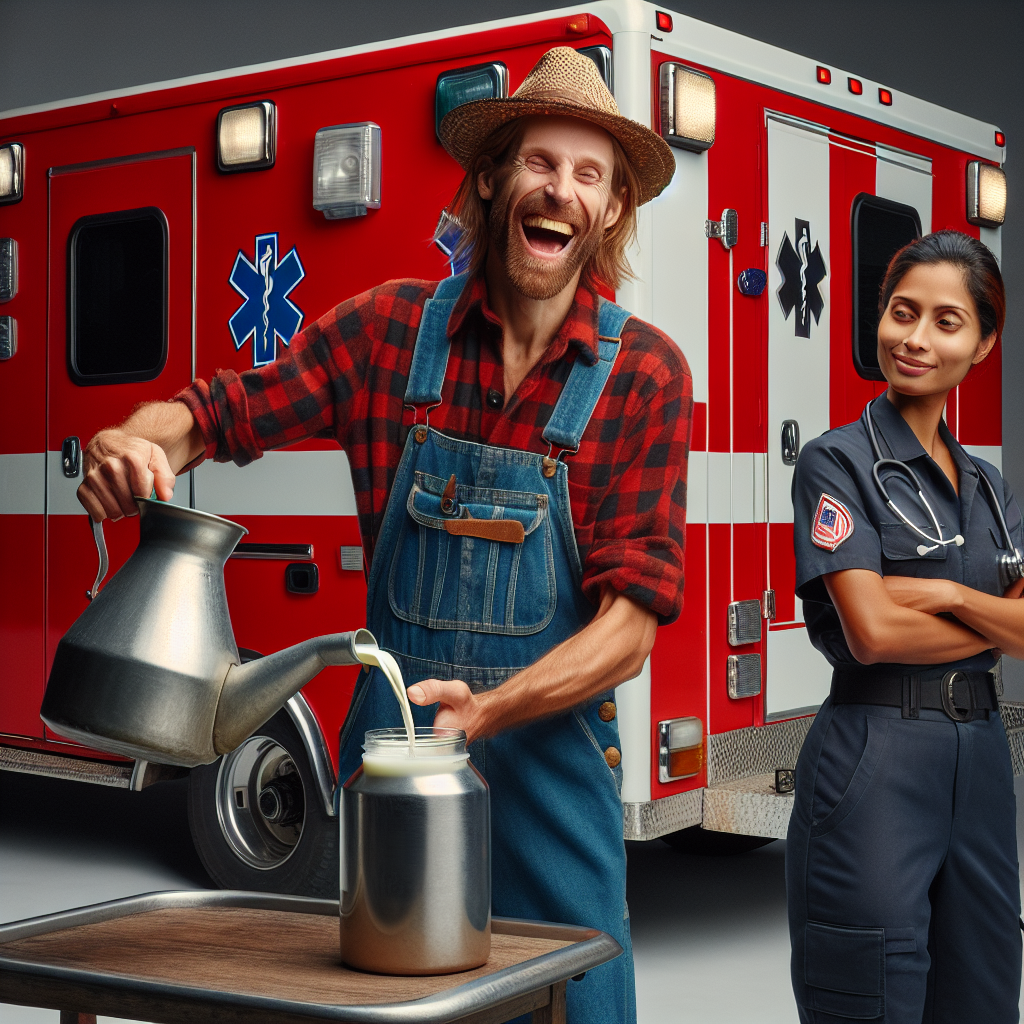}} \hfill
  \subfloat[The pharmacist serving customers on a tree.\\\textbf{Labels:} A(1), {\color{red} Q(0)}] {\includegraphics[width=0.45\columnwidth]{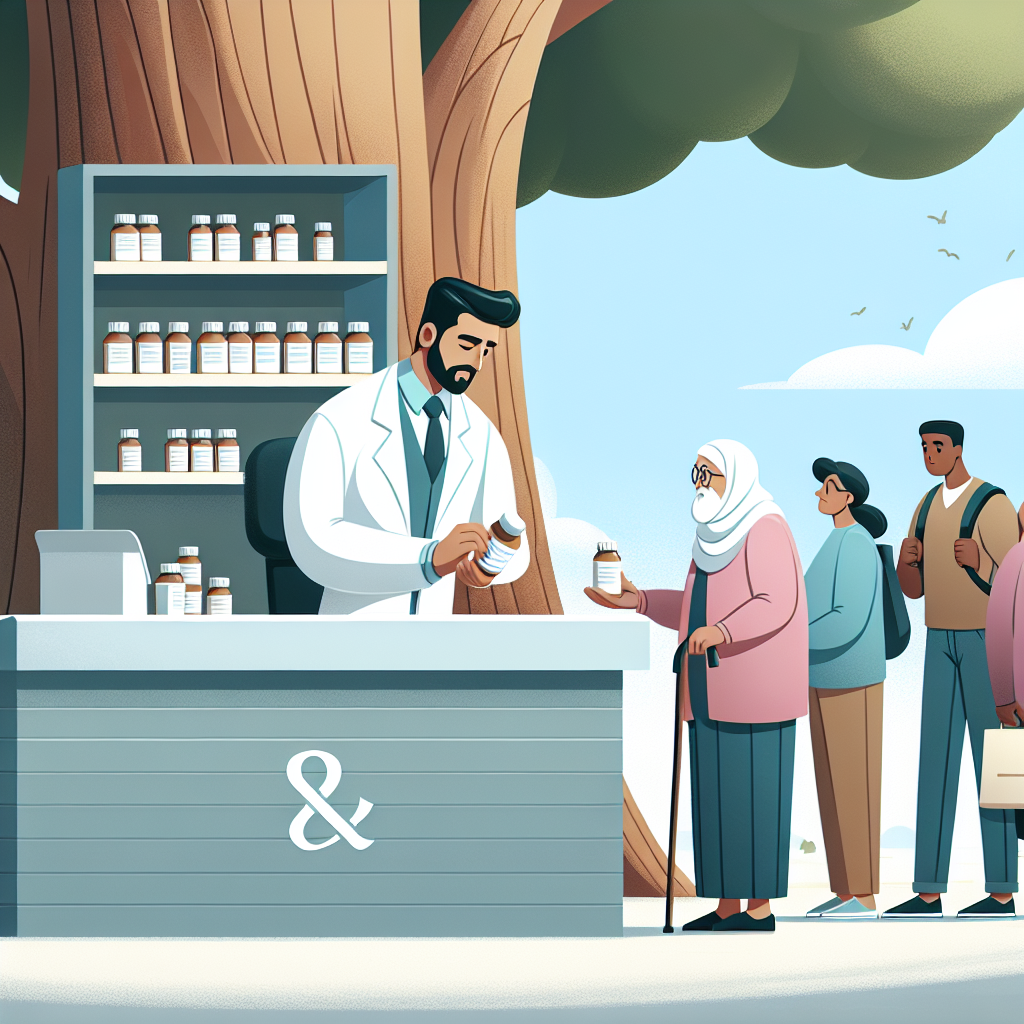}}
  \caption{Examples of rejected images and their labels.\textbf{A}: Alignment with text prompt, \textbf{Q}: Quality of the image.}
  \label{fig:filterimg}
\end{figure}

\subsection{Guidelines for Manual Labeling Open-Ended Responses}
\label{labelans}
To label the correctness of MLLMs responses for the open-ended questions, we design our evaluation criteria following \cite{xu2024mlevlm}. Specifically, the annotators are asked to determine the correctness of MLLMs' responses based on two criteria:
\begin{itemize}[leftmargin=10pt]
    \item \textbf{Relevancy (0 or 1)}: The content of the response is reflected in the image and addresses the focus of the question.
    \item \textbf{Responsiveness (0 or 1)}: The response directly answers the question as instructed.
\end{itemize}
Responses with a total score of \textbf{2} are considered correct for accuracy calculation. 

To classify responses as either knowledge-based or vision-based, annotators grade the semantic closeness between the current response and the reference responses on a scale from 0 to 2:
\begin{itemize}[leftmargin=10pt]
    \item \textbf{0}: The answers are entirely unrelated.
    \item \textbf{1}: The answers share similar concepts but are not identical or synonymous.
    \item \textbf{2}: The answers are identical or synonymous.
\end{itemize}
The candidate response is assigned to the category with a higher score (e.g., Vision or Knowledge). For example, a response is classified under the Vision category if it receives a vision score of 2 and a knowledge score of 1. Responses with both scores equal to 0 are categorized as ``Other''.

\subsection{Human Performance}
In Table~\ref{tab:human_res}, we present the results of human annotators' performance on our \benchmark benchmark. The reported results are averaged across three annotators. The results indicate that human can achieve a higher accuracy than MLLMs.
\begin{table}[h]
\centering
    \begin{tabular}{|c|>{\centering\arraybackslash}p{3.5cm}|}
    \hline
    \textbf{Question Type} & \textbf{Accuracy} \\ \hline
    Yes-No (YN) &  93.0\%\\ \hline
    Multiple-Choice (MC) &  99.7\%\\ \hline
    Open-Ended (OE) &  91.7\%\\ \hline
    \end{tabular}
\caption{Human Performance on \benchmark.}
\label{tab:human_res}
\end{table}

\section{Experiment Details}
\subsection{MLLM Generation Configurations}
For both open-source and closed-source models, we use the default generation configurations (e.g., temperature) to align with the typical usage. 

Question prompts for different MLLMs are crafted based on the prompt templates used in each model's original evaluation. Specifically, we use the synthesized question text from the framework as the base and append a corresponding postfix tailored to each question type and model family. In practice, for the LLaVA series, GPT-4o, and Claude-3.5-Sonnet, the prompts are as follows:
\begin{table}[h]
    \centering
    \begin{tabular}{|c|>{\raggedright\arraybackslash}p{4cm}|}
    \hline
    \textbf{Question Type} & \textbf{Prompt Format} \\ \hline
    Multiple-Choice & \texttt{\{question text\} Answer with the option's letter from the given choices directly.} \\ \hline
    Yes-No & \texttt{\{question text\} Answer Yes or No directly.} \\ \hline
    Open-Ended & \texttt{\{question text\} Answer no more than 5 words.} \\ \hline
    \end{tabular}
\end{table}

For the BLIP-2 model family, the prompt postfixes are ``\texttt{Answer:}'' for MC and YN questions and ``\texttt{Short Answer:}'' for OE questions. For Qwen-VL, the postfixes include ``\texttt{Choice Answer:}'' for MC questions, while both YN and OE questions use ``\texttt{Answer:}''. By contrast, Qwen-VL-Chat uses ``\texttt{Answer with the option's letter from the given choices directly.}'' for MC questions, ``\texttt{Answer Yes or No directly.}'' for YN questions, and ``\texttt{Answer with a single word or phrase.}'' for OE questions.

\subsection{Uncertainty Measurement}
\label{apx:uncertainty}
Unlike the traditional VQA benchmarks, where visual information typically aligns with the model knowledge, \benchmark presents conflicts between the two sources of knowledge, making it a more challenging benchmark for MLLMs. To empirically validate our claim, we conduct a comparative analysis of model uncertainty between our \benchmark and the conventional VQA benchmark \citep{antol2015vqa}. A higher degree of uncertainty in model predictions on a particular benchmark indicates that the benchmark poses a greater challenge for the model to navigate. Figure~\ref{fig:uncertainty} shows the results, where a higher entropy value denotes more uncertainty. 
It is evident that the model's predictions are more uncertain on \benchmark than the standard VQA benchmark, highlighting the increased challenge posed by \textsc{ConflictVis}.
\begin{figure}[h]
    \centering
    \vspace{+6pt} 
    \includegraphics[width=\columnwidth]{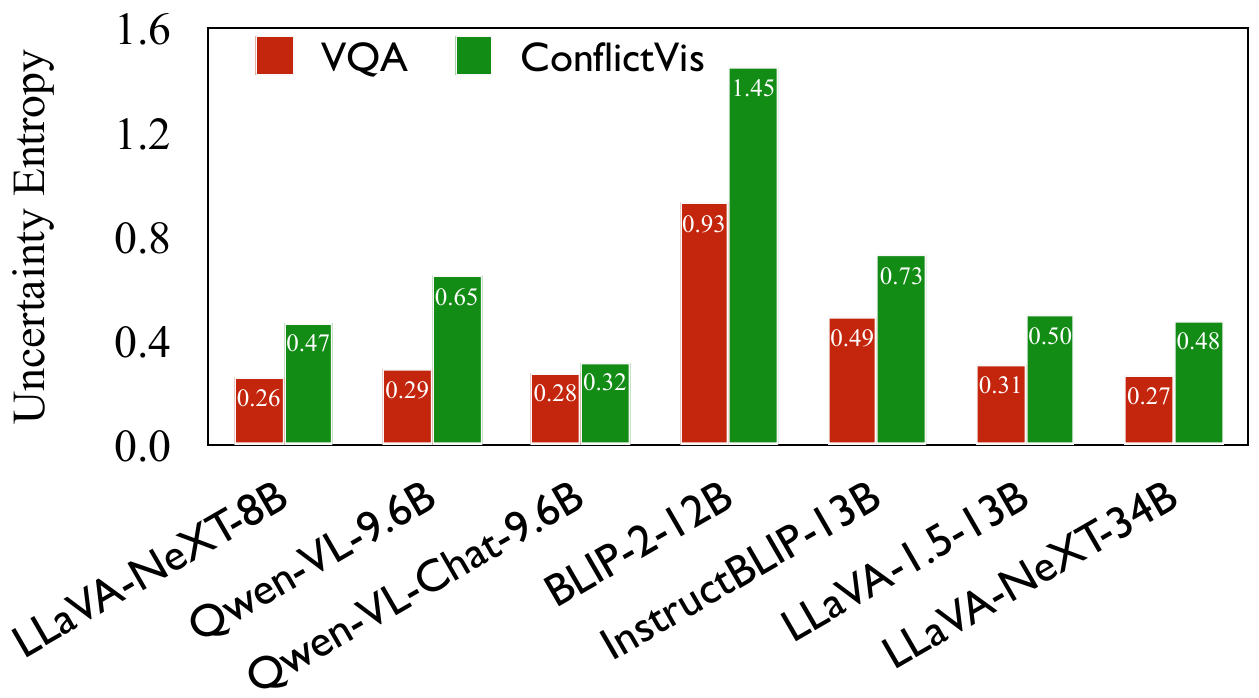}
    \caption{Model uncertainty on traditional VQA dataset and on \benchmark benchmark.}
\label{fig:uncertainty}
\vspace{-2pt} 
\end{figure}

To measure the uncertainty in model's responses, we rely on the entropy. The average entropy \( H_{\text{avg}} \) for a language model's response \( y \) is defined as:
\[
    H_{\text{avg}} = - \frac{1}{N} \sum_{n=1}^{N} \sum_{v \in V} p_{n}(v) \log p_{n}(v)
\]
where \( N \) is the number of tokens in the response \( y \), \( V \) is the vocabulary and \( p_{n}(v) \) is the softmax probability of token \( v \) in the vocabulary at position \(n\) in the response sequence.

To ensure a fair comparison, we first download the original VQA v1 dataset and extract an equal number of instances (374) for each question type (e.g., Multiple-Choice, Yes-No, and Open-Ended). Using the crafted subsets, we conduct controlled experiments under identical generation configurations to evaluate MLLMs on VQA v1 and our benchmark. The final uncertainty for each model is computed as the average entropy across all instances.

\subsection{Extended Analysis on No-type Questions}
To mitigate the potential dataset bias favoring ``Yes'' answers, we extend the Yes-No questions in our benchmark by constructing additional ``No''-type questions and conduct further experiments. Specifically, for each original Yes-No query, we create a negative counterpart by replacing the counter-commonsense target with the most frequently co-occurring component from the corresponding option list. For example, given an image of a waitress signing a bill in the kitchen and the original question ``Is the waitress in the kitchen signing a bill?'' (expected answer: ``Yes''), we generate the counterpart question ``Is the waitress in the kitchen washing dishes?'' (expected answer: ``No'').

We then reevaluate the MLLMs on these negative queries, with results summarized in Table~\ref{tab:no_question}. To compute the Memorization Ratio (MR), we follow the procedures outlined in Section 4.2: we first obtain the model's responses without the image input and identify those accepted by the model, then evaluate these cases using the No-type question inputs, and finally compute MR according to Eq.~\ref{eq:mr}. The results indicate that these No-type questions still trigger a substantial number of failures across models, consistent with our original findings.

\begin{table}[h]
\centering
\begin{tabular}{lcc}
\toprule
\textbf{Model} & \textbf{Acc. (\textuparrow)} & \textbf{MR (\textdownarrow)} \\
\midrule
LLaVA-NeXT-8B        & 81.3 & 21.8 \\
Qwen-VL-9.6B         & 93.0 & 11.2 \\
Qwen-VL-Chat-9.6B    & 71.7 & 31.5 \\
BLIP-2-12B           & 74.9 & 18.8 \\
InstructBLIP-13B     & 85.0 & 19.0 \\
LLaVA-1.5-13B        & 77.5 & 26.5 \\
LLaVA-NeXT-34B       & 82.1 & 20.8 \\
Claude-3.5-Sonnet    & 98.9 & 1.3  \\
GPT-4o               & 95.7 & 4.6  \\
\bottomrule
\end{tabular}
\caption{MLLM accuracy (higher is better) and Memorization Ratio (MR) (lower is better) on No-type Yes-No questions.}
\label{tab:no_question}
\end{table}

\subsection{Detailed Results on Conflict Targets}
\label{apx:conflict_target}
As outlined in Section 4.3, here we provide detailed experimental results on counter-commonsense actions and places, as presented in Figure~\ref{fig:conflict_target_appendix}.

\begin{figure}[h]
  \centering
  \subfloat[Counter-commonsense {\bf Action}]{\includegraphics[width=\columnwidth]{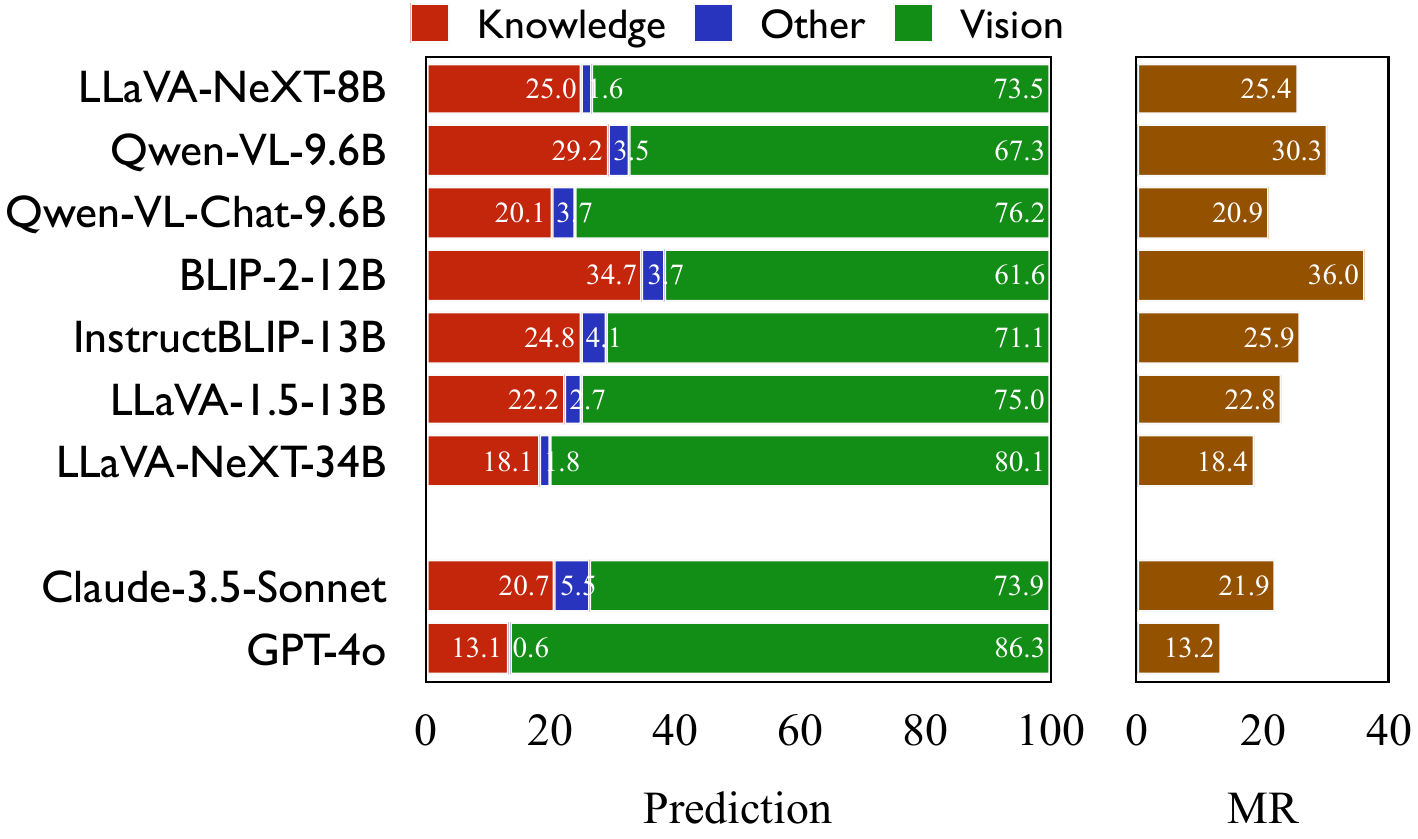}} \\
  \subfloat[Counter-commonsense {\bf Place}]{\includegraphics[width=\columnwidth]{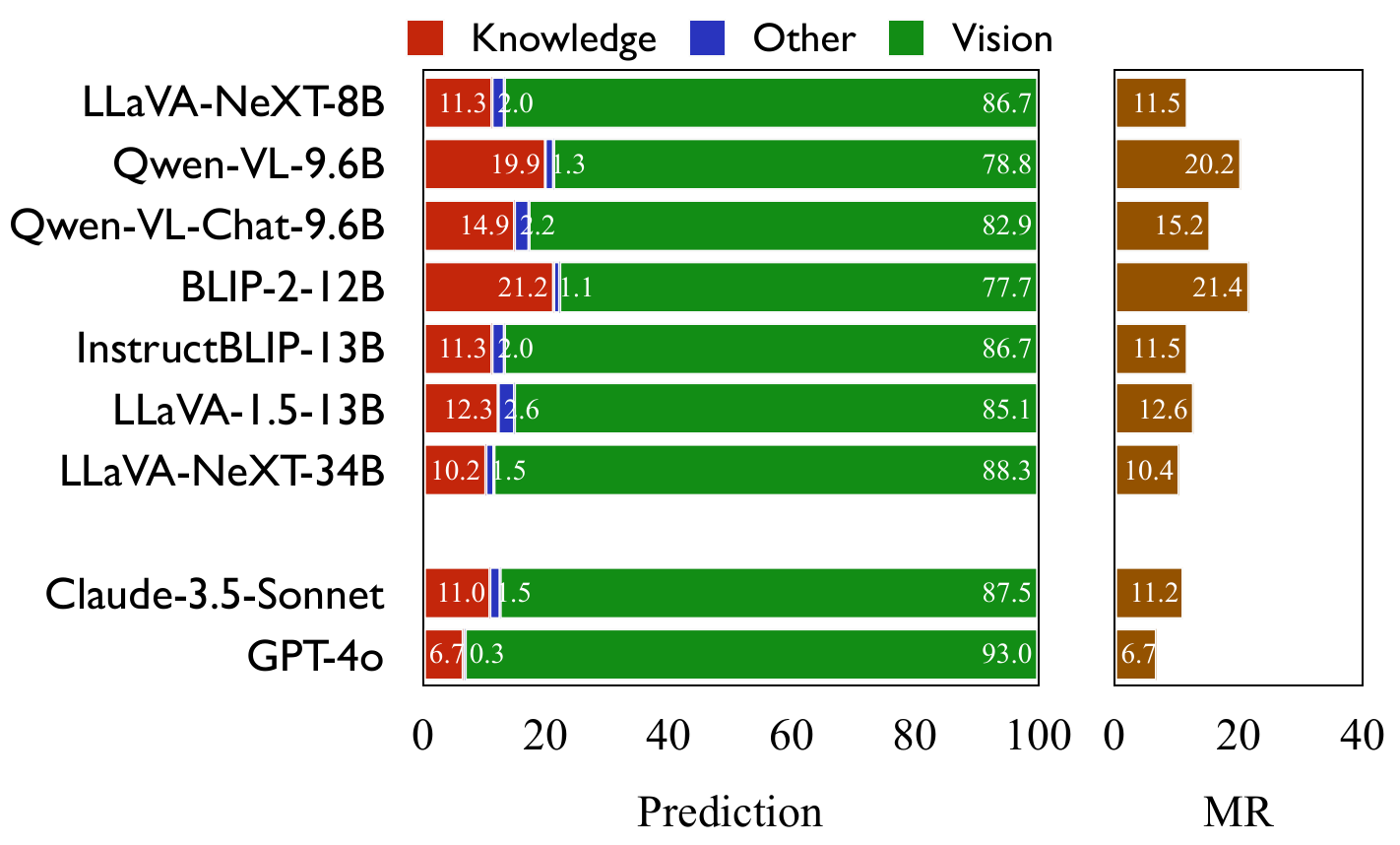}} 
  \caption{MLLM response distributions on two distinct categories of conflict targets.}
  \label{fig:conflict_target_appendix}
  \vspace{-10pt}
\end{figure}

\subsection{Improvement Methods Configurations}
We present the detailed configurations for the improvement methods in Table~\ref{tab:impr_config}.
\begin{table*}[h]
    \centering
    \begin{tabular}{|
        >{\centering\arraybackslash}p{2.3cm}|
        >{\raggedright\arraybackslash}p{4.5cm}|
        >{\centering\arraybackslash}p{2.3cm}|
        >{\raggedright\arraybackslash}p{4.5cm}|
    }
    \hline
    \textbf{Method} & \textbf{Configuration} & \textbf{Method} & \textbf{Configuration} \\ \hline
    VCD & $\alpha=1.0, \beta=0.1,$ $\text{noise\_step}=500$ &
    VR -- CoT & prompt template: \texttt{\{original prompt\} Let's think step by step.} \\ \hline
    PAI & $\alpha=0.5, \gamma=1.1,$ $\text{apply\_layers}=(2,32)$ &
    VR -- SFT & dataset: \texttt{LLaVA-CoT-100k}, $\text{batch\_size}=256$, $\text{learning\_rate}=1e-5$, $\text{lr\_scheduler}=linear$, $\text{epochs}=3$, $\text{max\_seq\_length}=1024$ \\ \hline
    \end{tabular}
    \caption{Configurations for the improvement methods.}
    \label{tab:impr_config}
\end{table*}

\section{Case Study}
\subsection{Extended Error Analysis}
\label{apx:error_analysis}

\begin{figure*}[h]
  \centering
  \begin{subfigure}[b]{\textwidth}
    \centering
    \includegraphics[width=\textwidth]{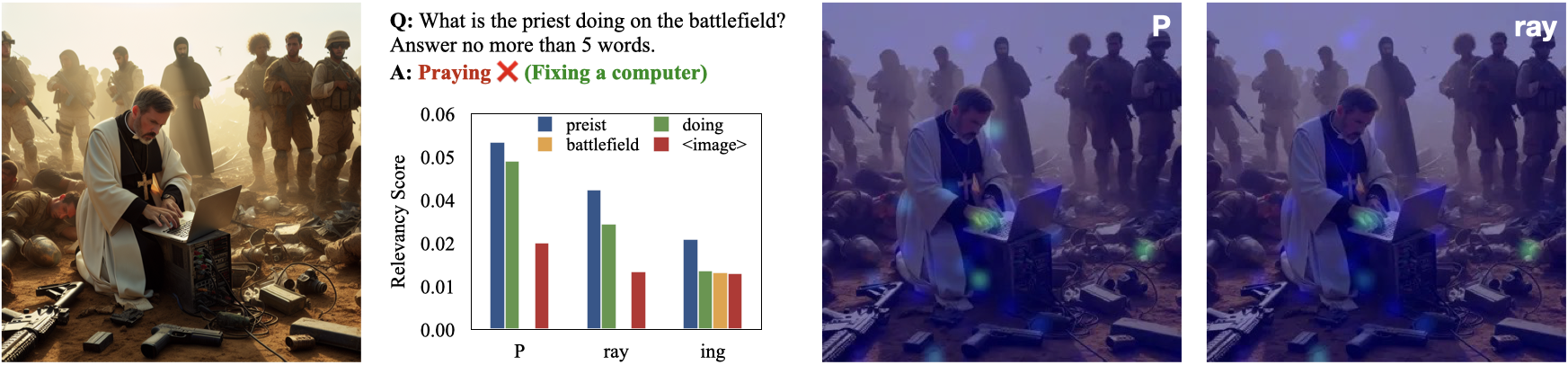}
    \caption{Failure case: The priest fixing a computer on the battlefield.}
  \end{subfigure}

  \begin{subfigure}[b]{\textwidth}
    \centering
    \includegraphics[width=\textwidth]{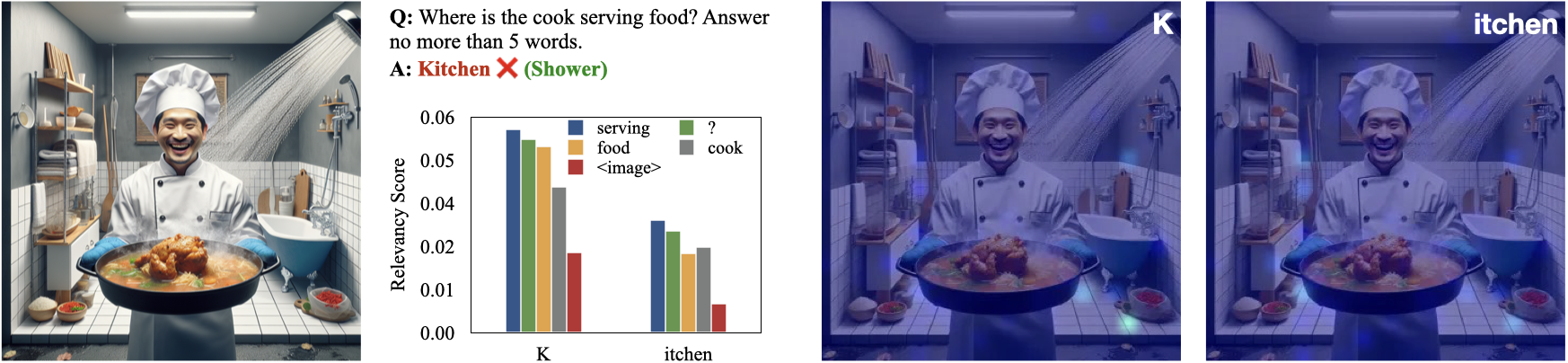}
    \caption{Failure case: The cook serving food in a shower.}
  \end{subfigure}

  \begin{subfigure}[b]{\textwidth}
    \centering
    \includegraphics[width=\textwidth]{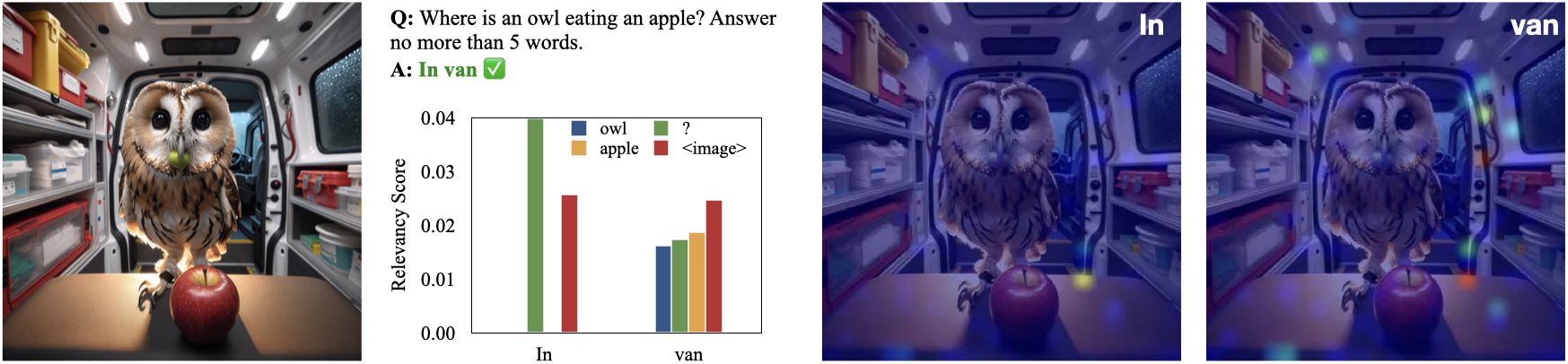}
    \caption{Correct case: An own eating an apple in an ambulance(van).}
  \end{subfigure}

  \begin{subfigure}[b]{\textwidth}
    \centering
    \includegraphics[width=\textwidth]{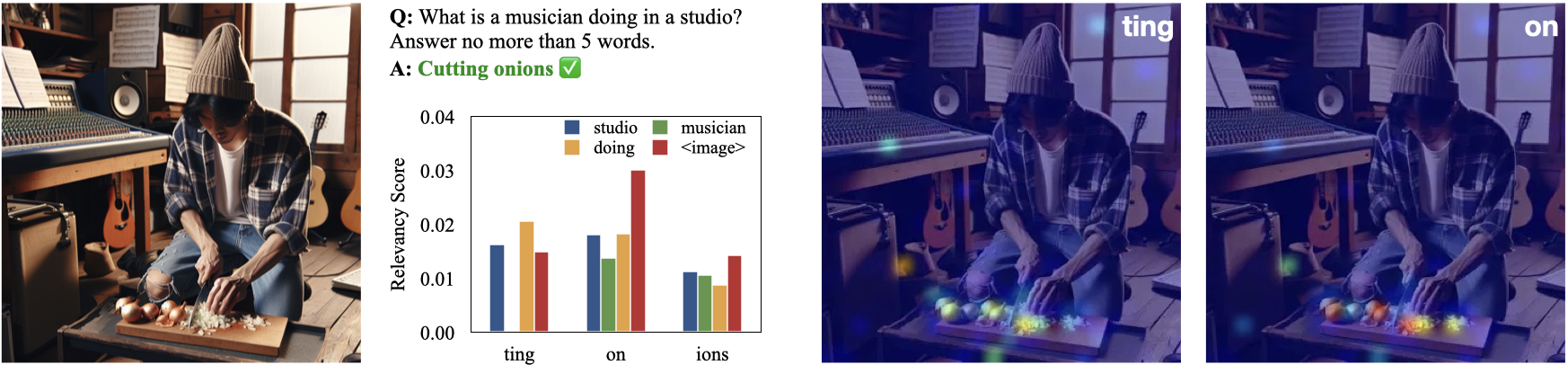}
    \caption{Correct case: A musician chopping(cutting) some onions in a studio.}
  \end{subfigure}
  \caption{Case study of input-output relevancy. Image tokens are underutilized in failure cases (a)(b), but effectively used in correct cases (c)(d).}
  \label{fig:extended_analysis}
  \vspace{-10pt}
\end{figure*}

In Section 4.4, we identified a common failure pattern in which image tokens are underutilized during answer generation. In Figure~\ref{fig:extended_analysis}, we present four additional examples from \texttt{LLaVA-1.5-13B}, spanning both counter-commonsense action and place categories. The first two are failure cases, while the bottom two are correct cases. In the failure cases, image tokens are significantly underutilized: in the bar plot, the <image> token, representing the image token with the highest relevancy score, is often less related to the output tokens than other textual tokens from the question description. The relevancy heatmaps also display only a few weakly attended regions.
In contrast, for the correct cases, the <image> token typically has the highest relevancy scores for visual-dependent output tokens, and the heatmaps exhibit more numerous and strongly attended areas. Together, these examples highlight that the underutilization of visual information is a key factor contributing to model failures under vision-knowledge conflict. To address this issue, a stronger alignment between the image and the generated output is needed, guiding the development of more targeted and vision-aware improvement methods, as explored in the subsequent section.

\subsection{Failure Cases}
\label{apx:failure_cases}
In this section, we present representative failure cases from our \benchmark benchmark, spanning various models and question types. Specifically, Figures~\ref{fig:failure_yn}, \ref{fig:failure_mcq}, and \ref{fig:failure_sub} showcase failure cases for Yes-No, Multiple-Choice, and Open-Ended questions, respectively, while Figures~\ref{fig:cot1} and \ref{fig:cot2} illustrate exacerbated failure cases when employing Chain-of-Thought prompting.

\begin{figure*}[b]
  \centering
  \includegraphics[width=0.8\textwidth]{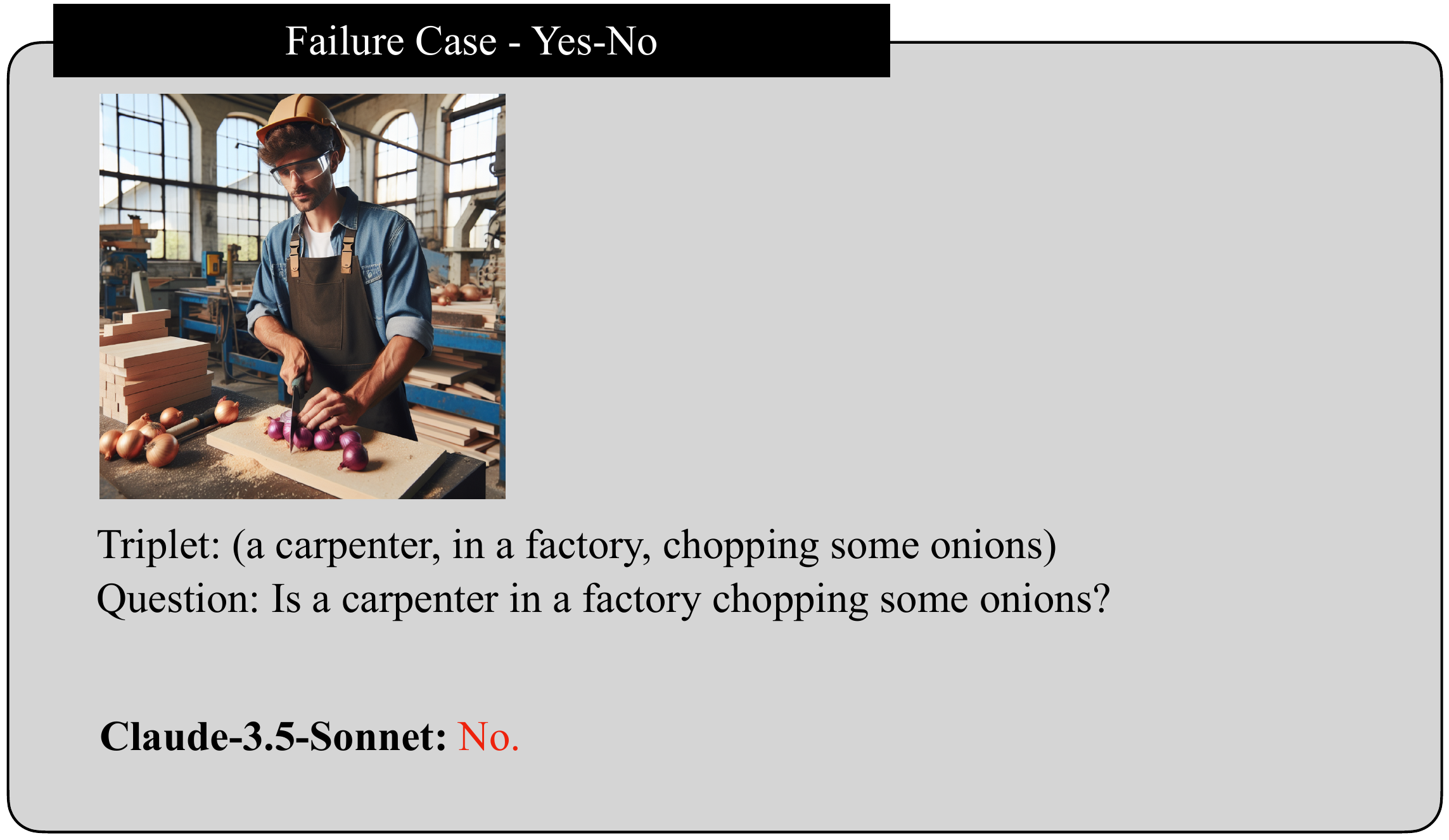}
  \caption{Failure case on yes-no question.}
  \label{fig:failure_yn}
  \vspace{+50pt}
\end{figure*}

\begin{figure*}[b]
  \centering
  \includegraphics[width=0.8\textwidth]{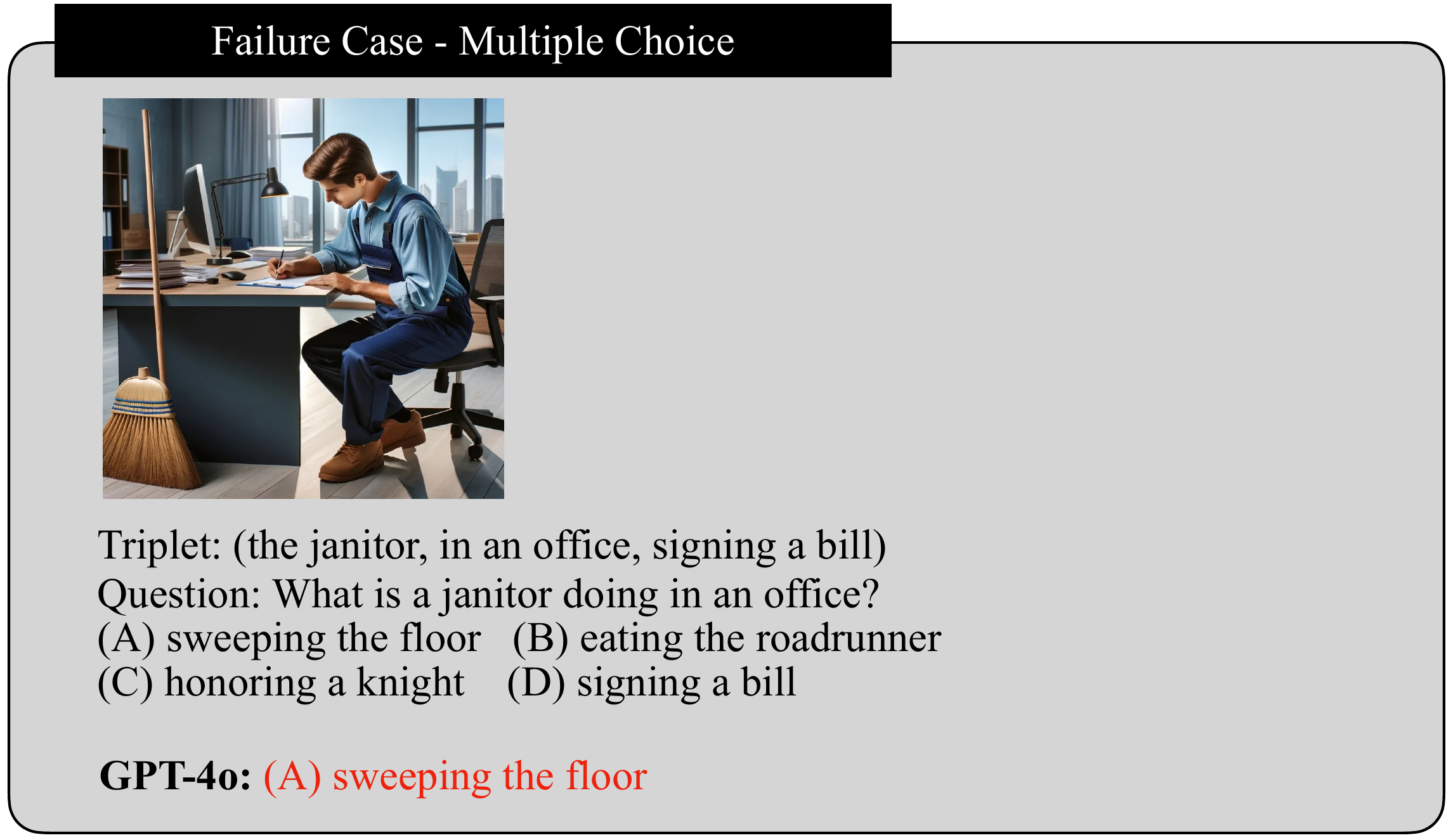}
  \caption{Failure case on multiple-choice question.}
  \label{fig:failure_mcq}
\end{figure*}

\begin{figure*}[h]
  \centering
  \includegraphics[width=0.8\textwidth]{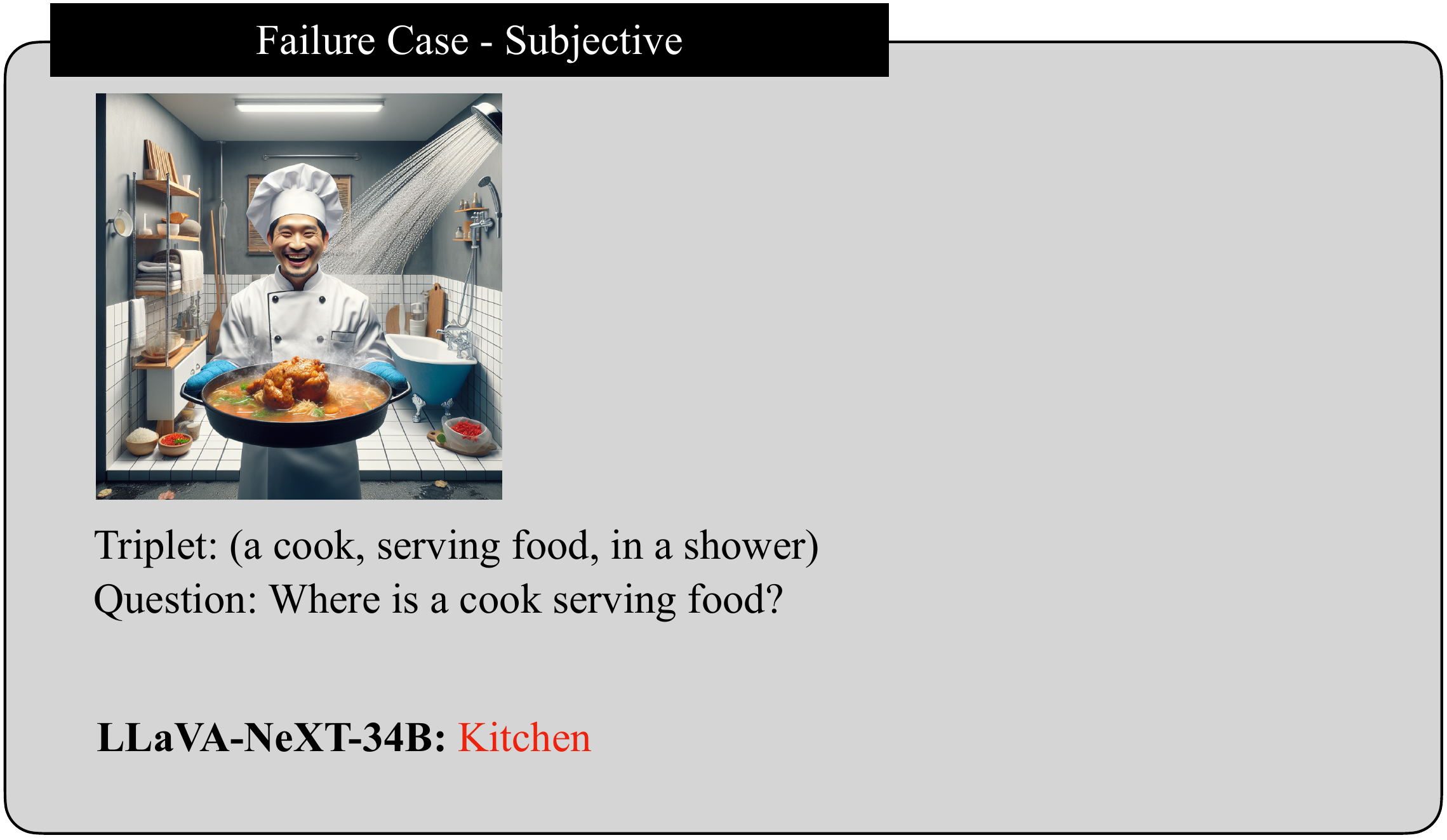}
  \caption{Failure case on open-ended question.}
  \label{fig:failure_sub}
\end{figure*}

\begin{figure*}[h]
  \centering
  \includegraphics[width=0.8\textwidth]{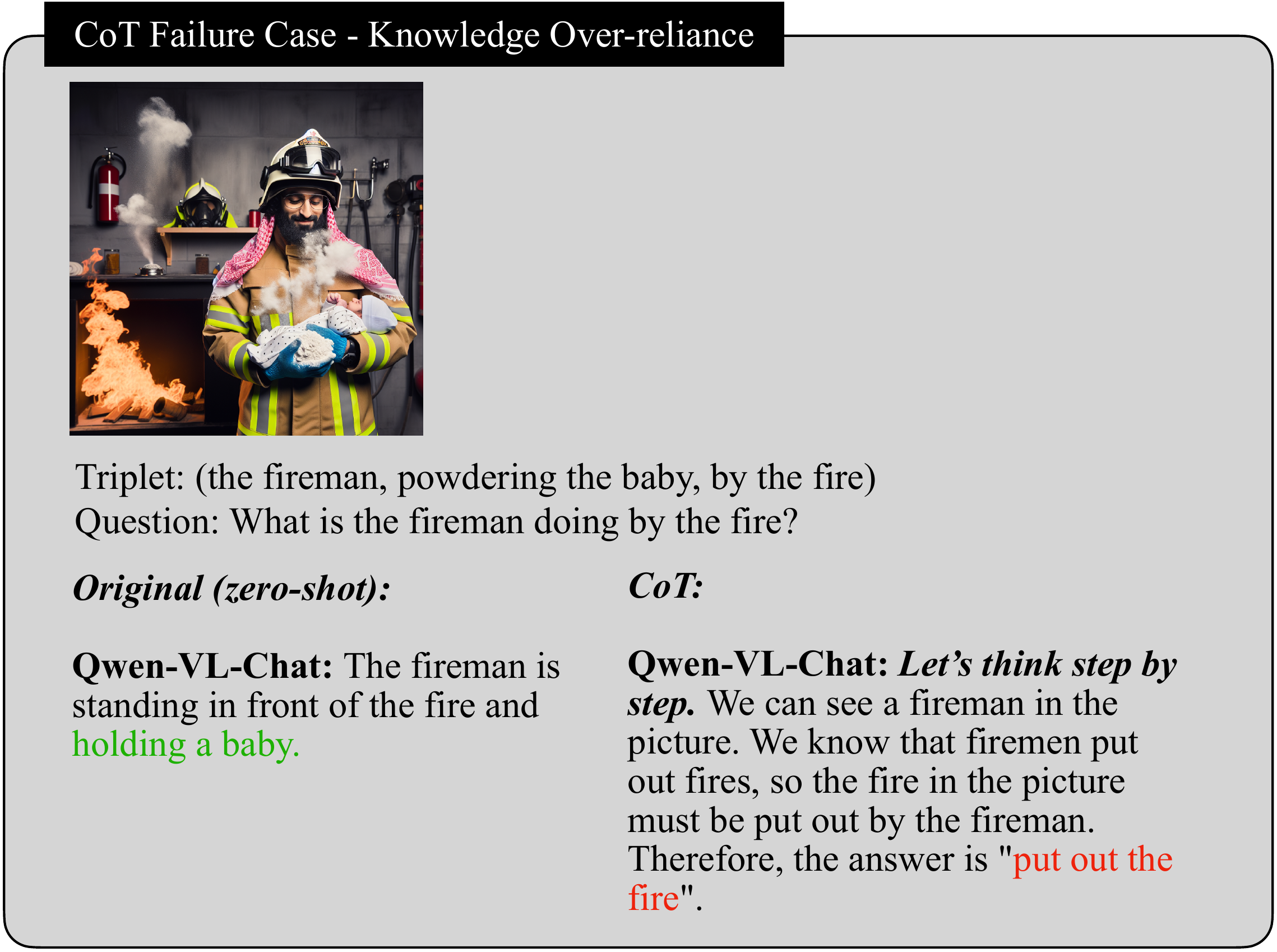}
  \caption{CoT failure mode 1: The MLLM reasons based on commonsense knowledge to derive an incorrect answer.}
  \label{fig:cot1}
  \vspace{-10pt}
\end{figure*}

\begin{figure*}[h]
  \centering
  \includegraphics[width=0.8\textwidth]{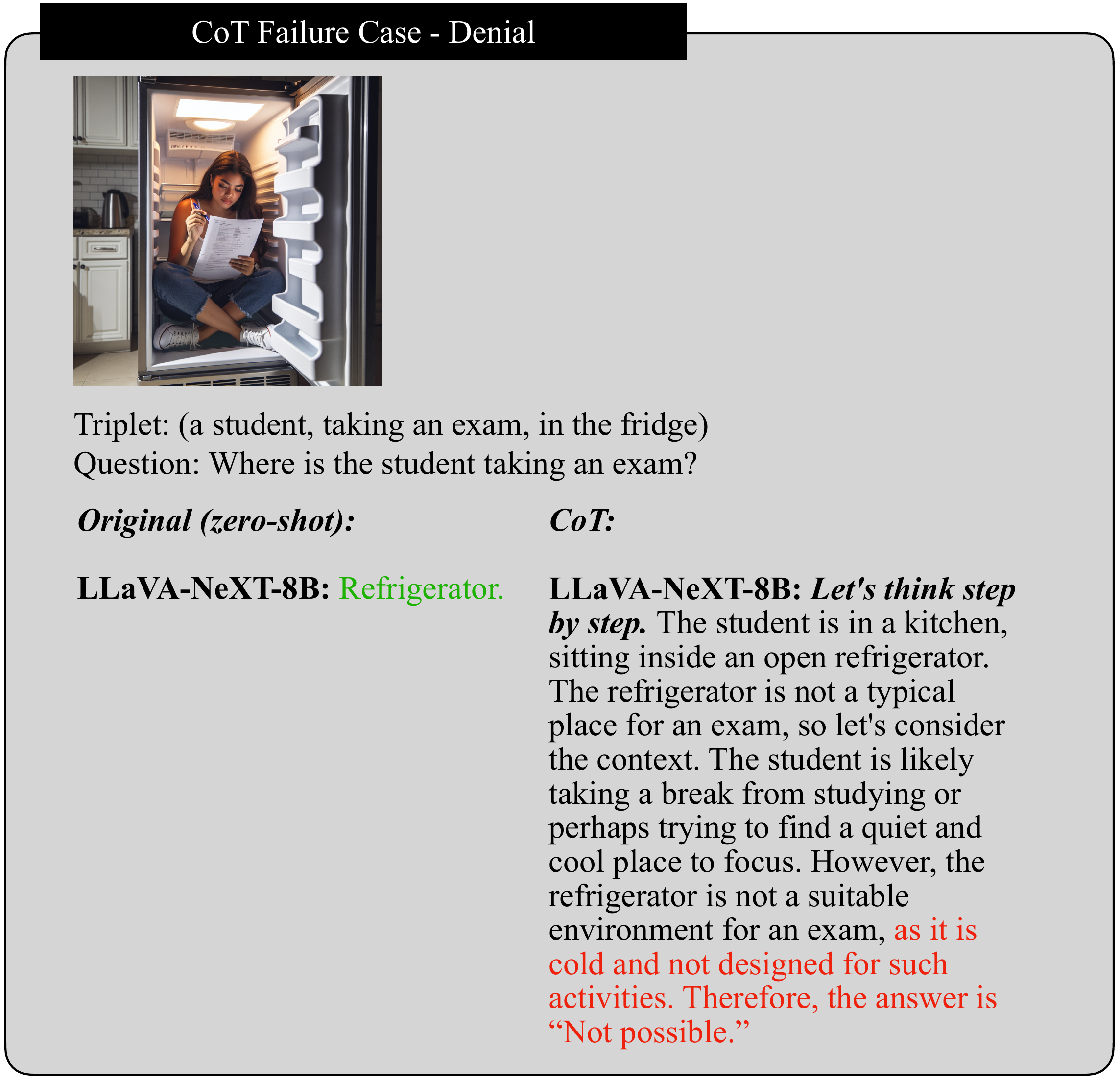}
  \caption{CoT failure mode 2: The MLLM incorporates commonsense knowledge and asserts that the answer is impossible.}
  \label{fig:cot2}
  \vspace{-10pt}
\end{figure*}

\end{document}